\pdfoutput=1
\documentclass[11pt]{article}
\usepackage{samaya}
\titleformat{\paragraph}[runin]{\normalfont\bfseries\color{SamayaInk}}{}{0pt}{}[.]
\usepackage[numbers,sort&compress]{natbib}
\usepackage{array,tabularx,multirow}
\usepackage{tikz}
\usetikzlibrary{arrows.meta,positioning,fit,backgrounds,calc,shapes.geometric,
                decorations.pathreplacing,patterns}
\usepackage{pgfplots}   
\pgfplotsset{compat=1.18}

\definecolor{figAcq}{HTML}{2A78D6}   
\definecolor{figFil}{HTML}{EB6834}   
\definecolor{figCon}{HTML}{1BAF7A}   
\definecolor{figGood}{HTML}{0CA30C}  
\definecolor{figBad}{HTML}{D03B3B}   
\definecolor{figInk}{HTML}{0B0B0B}   
\definecolor{figInk2}{HTML}{52514E}  
\definecolor{figMute}{HTML}{8A8984}  
\definecolor{figSurf}{HTML}{F0EFEC}  
\definecolor{figSurf2}{HTML}{FAFAF8} 

\definecolor{figAcqBg}{HTML}{E4EEFB}
\definecolor{figFilBg}{HTML}{FCE9E0}
\definecolor{figConBg}{HTML}{E0F5EE}
\definecolor{figGoodBg}{HTML}{E2F4E2}
\definecolor{figBadBg}{HTML}{F9E5E5}

\tikzset{
  figfont/.style   = {font=\sffamily\fontsize{7.4}{9}\selectfont, text=figInk},
  figfontsm/.style = {font=\sffamily\fontsize{6.6}{8}\selectfont, text=figInk2},
  figfontlg/.style = {font=\sffamily\bfseries\fontsize{8}{10}\selectfont, text=figInk},
  figmono/.style   = {font=\ttfamily\fontsize{7}{8.5}\selectfont, text=figInk},
  fbox/.style = {figfont, draw=figMute, line width=0.5pt, rounded corners=1.6pt,
                 fill=figSurf2, inner xsep=5pt, inner ysep=3.6pt,
                 align=center, minimum height=15pt},
  facq/.style = {fbox, draw=figAcq, fill=figAcqBg},
  ffil/.style = {fbox, draw=figFil, fill=figFilBg},
  fcon/.style = {fbox, draw=figCon, fill=figConBg},
  fgood/.style= {fbox, draw=figGood, fill=figGoodBg},
  fbad/.style = {fbox, draw=figBad,  fill=figBadBg},
  fplain/.style={fbox, draw=figMute, fill=figSurf},
  fghost/.style={fbox, draw=figMute, fill=none, dash pattern=on 2pt off 1.6pt},
  fband/.style = {figfontlg, inner sep=2pt, align=center},
  farr/.style  = {-{Stealth[length=4pt,width=3pt]}, draw=figInk2, line width=0.6pt},
  farrm/.style = {-{Stealth[length=4pt,width=3pt]}, draw=figMute, line width=0.55pt},
  farrd/.style = {-{Stealth[length=4pt,width=3pt]}, draw=figMute, line width=0.55pt,
                  dash pattern=on 2pt off 1.6pt},
  fline/.style = {draw=figMute, line width=0.5pt},
  flab/.style  = {figfontsm, fill=white, inner sep=1.4pt},
  faxis/.style = {draw=figMute, line width=0.5pt},
  ftick/.style = {figfontsm},
  fgroup/.style= {draw=figMute, line width=0.5pt, rounded corners=2.5pt,
                  dash pattern=on 2pt off 1.6pt, inner sep=6pt},
}

\newcommand{\figblock}[1]{%
  \par\addvspace{7pt}%
  {\centering#1\par}%
  \addvspace{5pt}%
}

\newcommand{\figfontsmtxt}{\fontsize{6.6}{8}\selectfont\color{figInk2}}
\newcommand{\figfontlgtxt}{\sffamily\bfseries\fontsize{8}{10}\selectfont\color{figInk}}

\newcommand{\cc}[1]{{\small\ttfamily\def\_{\textunderscore\allowbreak}#1}}
\newcommand{\figcap}[2]{{\small\color{SamayaInkSoft}{\sffamily\bfseries\color{SamayaInk}Figure~#1.}\ #2\par}}

\title{Hindsight Structured Advice Distillation: Steering Executor Agents with Advisors Trained on Their Failures and Task Characteristics}
\author{O. Ozan Koyluoglu\\[4pt]
\samayaaffil{Samaya AI}}
\date{}
\samayacorrespondence{\samayaemail{ozan@samaya.ai}}
\samayapreprint{Samaya AI Technical Report}

\usepackage{float}
\begin{document}

\begin{samayatitle}
Language-model agents are used for specialized work such as financial research, healthcare and law. Such agents are often capable enough for a task but still miss its details. In finance, for example, these details include which filing reports a figure or how a company discloses a metric. Retraining the agent to learn these details is expensive, and often impossible when the agent is a closed or shared system. Writing the missing know-how into the agent's prompt makes the prompt long and identical for every input, although different inputs need different guidance. We propose to learn to generate this guidance instead. A small \emph{advisor} model reads each input and writes structured advice, which is appended to the prompt of a frozen executor. The advisor is trained with hindsight, following work on distillation with privileged information. On a training pool with graded answers, we analyse the executor's failures, run it under several strategies to see which one worked on each input, and let a teacher that sees the grades and the traces write the advice that would have helped. The advisor then learns, by supervised fine-tuning, to write that advice from the input alone. Because the student is a separate small model rather than the executor, it is cheap to train and adds little latency. We call this approach \emph{hindsight structured advice distillation} (HSAD). On FrontierFinance, a public benchmark of expert-written financial research queries, an 8B advisor raises the qualification rate of Samaya's proprietary research agent, which already leads the benchmark's leaderboard, from 52.9\% to 56.6\%. This is above the agent's own high-effort configuration, at 37\% lower cost per query. We study the generalization of the proposed approach, and in particular show that HSAD also improves results on WebShop, BIRD text-to-SQL and ExpertLongBench.
\end{samayatitle}
\suppressfloats[t]

\section{Introduction}

Language-model agents now browse websites, query databases, call tools and write research reports \cite{arXiv221003629,arXiv220701206,arXiv250902547}. In many deployments, the model at the core of the agent cannot be changed. It may be a closed model served through an API, a production agent shared by many teams, or a small open model that is cheap to serve and costly to re-validate after every change. Improving such a system therefore usually means changing what it is told rather than its weights.

What the agent needs to be told is often know-how about the task rather than general ability. For example, a financial research agent may be asked for a company's quarterly revenue by product category when the company publishes only each category's percentage share of sales, not its revenue in dollars. An agent that does not know this may fill the table with its own estimates, when the better answer reports the percentages and states that dollar figures are not disclosed. A shopping agent may buy the right shirt in the wrong size, because it did not work out which option buttons the instruction required. In both cases, the agent has the knowledge and the tools it needs. It is missing a detail of how this kind of task should be done, and that detail differs from one input to the next.

\begin{figure}[t]
\figblock{\resizebox{\linewidth}{!}{
\begin{tikzpicture}[figfont, node distance=4mm]

\node[fplain, text width=2.0cm] (pool) at (0,0) {\textbf{train pool}\\[1pt]{\figfontsmtxt questions $x$ + gold $y^\star$}};
\node[facq, text width=2.3cm, right=7mm of pool] (arms) {\textbf{frozen executor}\\[1pt]{\figfontsmtxt runs every arm\\$a_1,\dots,a_K$ on each $x$}};
\node[fgood, text width=2.3cm, right=7mm of arms] (win) {\textbf{hindsight label}\\[1pt]{\figfontsmtxt graded against $y^\star$;\\winner $a^\star(x)$, traces}};
\node[ffil, text width=2.4cm, right=7mm of win] (teach) {\textbf{teacher}\\[1pt]{\figfontsmtxt writes advice $z^\star$\\from $x$, $y^\star$, traces}};
\node[fcon, text width=2.1cm, right=7mm of teach] (sft) {\textbf{advisor SFT}\\[1pt]{\figfontsmtxt LoRA on\\$(x \mapsto z^\star)$}};

\draw[farr] (pool) -- (arms);
\draw[farr] (arms) -- (win);
\draw[farr] (win) -- (teach);
\draw[farr] (teach) -- (sft);

\node[ffil, text width=4.6cm, above=6mm of $(arms.north)!0.5!(win.north)$] (mine)
     {\textbf{error analysis on base-executor failures}\\[1pt]{\figfontsmtxt mine arms (strategies) + directive taxonomy}};
\draw[farrm] ($(pool.north)$) |- (mine.west);
\draw[farr] (mine.south -| arms.north) -- (arms.north);
\draw[farrd] (mine.east) -| node[flab, pos=0.25]{closed vocabularies} (teach.north);

\node[fband, anchor=south west] at ($(pool.north west)+(0,1.55)$) {TRAINING TIME \,{\figfontsmtxt(gold visible)}};

\node[fplain, text width=2.0cm, below=17mm of pool] (q) {\textbf{new question}\\[1pt]{\figfontsmtxt $x$ only, no gold}};
\node[fcon, text width=2.3cm, right=7mm of q] (adv) {\textbf{trained advisor}\\[1pt]{\figfontsmtxt small LM, emits\\structured JSON $\hat z$}};
\node[fplain, text width=2.3cm, right=7mm of adv] (rend) {\textbf{render}\\[1pt]{\figfontsmtxt fields $\to$ text\\appended to prompt}};
\node[facq, text width=2.4cm, right=7mm of rend] (exe) {\textbf{same frozen executor}\\[1pt]{\figfontsmtxt base prompt + advice}};
\node[fgood, text width=2.1cm, right=7mm of exe] (ans) {\textbf{answer}\\[1pt]{\figfontsmtxt compared vs.\ no advice}};
\draw[farr] (q) -- (adv);
\draw[farr] (adv) -- (rend);
\draw[farr] (rend) -- (exe);
\draw[farr] (exe) -- (ans);
\draw[farrd] (sft.south) -- ++(0,-0.55) -| node[flab, pos=0.2]{weights} (adv.north);

\node[fband, anchor=south west] at ($(q.north west)+(0,0.22)$) {TEST TIME};
\end{tikzpicture}}}
\figcap{1}{HSAD. At training time (top), error analysis on the frozen executor's failures yields closed vocabularies of arms and directives. The executor is then run under every arm, and each input is labelled with the arm that won against the gold answer. A teacher that sees the gold answer and the traces writes a structured advice target, and a small advisor is fine-tuned to map the input to that target. At test time (bottom), the advisor sees only the input, and its advice is rendered as text and appended to the unchanged executor prompt.}
\end{figure}

There are two usual ways to supply such know-how. The first is to train the agent. Reinforcement learning (RL) is now a standard way to train language agents \cite{arXiv250902547}, but it is expensive when every rollout takes minutes and the reward is a rubric score, and it is not an option when the agent's weights are fixed. The second is to write the know-how into the agent's prompt, and a large body of work optimizes such prompts automatically \cite{arXiv221101910,arXiv230903409,arXiv231003714,arXiv240607496,arXiv250719457,arXiv251004618}. However, most of these methods produce one prompt or context for a whole task. The prompt grows long as it covers more cases, and much of it applies to only some inputs. Per-input steering also exists, for example a small policy model that writes a hint for each input \cite{arXiv230211520}, or an advisor model trained with RL to write per-instance advice for a black-box model \cite{arXiv251002453}. These methods still leave open what the advice should contain and where its training signal should come from.

We argue that hindsight can supply that signal. After the fact, on a training input with a graded answer, it is often easy to see what would have helped. The shopping episode above, for instance, shows that the agent needed to be told which size button to click. Hindsight relabeling has a long history in RL \cite{arXiv170701495}, in instruction following \cite{arXiv230205206}, and in reasoning \cite{arXiv220314465}. It is also the core of learning with privileged information, where a teacher that sees information unavailable at test time trains a student that does not \cite{arXiv151103643,arXiv241005434,arXiv260806065}. Recent work in this line distills the privileged teacher into the agent itself. We distill it into a separate small advisor instead, which leaves the agent unchanged and is cheap to train and to run. With this approach, the student must be able to infer, from its own inputs, what the teacher inferred with the privileged information, or the transfer fails \cite{arXiv241005434,arXiv260511182}.

We call the resulting approach \emph{hindsight structured advice distillation} (HSAD), shown in Fig.~1. At test time, we have a small trained advisor in front of a frozen executor (similar to \cite{arXiv230211520, arXiv251002453}), and we train the advisor by hindsight distillation. The training targets come from hindsight, what is distilled into the student model is advice, and the advisor writes that advice as a structured output (JSON) whose fields take values from closed vocabularies. First, we analyse the frozen executor's failures on a training pool and mine two vocabularies from them: a small set of \emph{arms}, which are whole-response strategies appended to the executor's base prompt, and a two-level \emph{directive taxonomy} of error areas and specific fixes. Second, we run the executor under every arm on every training input, grade the answers, and record which arm won. Third, a teacher writes the advice that would have helped most. The teacher combines an LLM, which sees the input, the grading, every arm's answer with its score and tool usage, the winning trace, and the closed vocabularies to choose from, with fixed rules that fill the fields that can be read off the runs or the input. Fourth, we train a small advisor by supervised fine-tuning (SFT) to write that advice from the input alone. At test time, this advice is rendered as text and appended to the executor prompt, where it acts as guidance that the executor is free to follow or ignore.

Our main setting is FrontierFinance, a public benchmark of 220 expert-written financial research queries graded against more than eleven thousand source-attributed rubric items \cite{arXiv260811683}. Its report finds that the harness (tools, prompts and so on) and the executor model strongly shape quality, and that Samaya's proprietary research agent leads all evaluated systems. We take that agent as the frozen executor and ask whether a small advisor can still improve it. On the public leaderboard, the advised agent reaches a 56.6\% qualification rate, against 52.9\% without advice and 56.0\% for the agent's own high-effort configuration, at 37\% lower cost per query than that configuration. We then apply HSAD with a small frozen Qwen3-4B executor \cite{arXiv250509388} in three further settings: WebShop \cite{arXiv220701206}, a simulated shopping environment; BIRD text-to-SQL \cite{arXiv230503111}; and ExpertLongBench \cite{arXiv250601241}, a set of expert-level long-form generation tasks. On WebShop, the score rises from 54.6 to 63.4 and the success rate increases from 3.8\% to 23.2\% on 500 held-out samples. On BIRD and ExpertLongBench, the gains are +3.1 execution-accuracy points and +4.9 checklist-coverage points, respectively. These results suggest that the approach carries over to other executors and domains.

These four settings also show how the advice helps. Part of its value is insight learned from the executor's traces. The vocabulary comes from error analysis of how this executor fails, and the labels come from which strategy worked on which input. On FrontierFinance, the directives mined in this way are one of the most important fields of the advice. The other part is instance-specific information that the advisor can supply and the executor tends to miss. On WebShop, copying the option values from the task into the advice as exact click targets improves performance. How much the advice helps therefore depends on the task, the training data and the executor. In our experiments, how the hindsight labels were defined had a clear effect on the trained advisor.

\paragraph{Contributions} (1) We propose HSAD, a supervised approach that learns from a frozen executor's traces, through error analysis and hindsight labels, and distills what it learns into a small advisor that writes structured, input-specific guidance. (2) We make explicit where each training label comes from, and show that both the judgments of the teacher's LLM and the labels its rules compute from the runs are useful. (3) We distinguish three reference points for how much advice can help. The first is the best strategy for each input as recorded during collection (best-arm oracle), the second is the score when the teacher's advice, or only its arm, is given to the executor in a new run, and the third is the score with the trained advisor's advice (the student's performance). (4) We show that a small advisor lifts a proprietary, leaderboard-leading research agent above its own high-effort configuration on the public FrontierFinance benchmark, and we report gains in three further settings with a small open executor.

\section{Related Work}

\paragraph{Steering a frozen model with a second model} Several lines of work keep a large model fixed and train a smaller one to influence it. Directional stimulus prompting trains a small policy model, first with SFT and then with RL, to write an instance-specific hint that is added to a black-box model's prompt \cite{arXiv230211520}. Proxy tuning steers a large model at decoding time by adding the logit difference between a small tuned model and its untuned version \cite{arXiv240108565}, which requires access to the large model's output distribution. Planner--executor splits divide an agent into a model that plans and a model that acts, either within a multi-LLM tool-use agent \cite{arXiv240107324} or with a separately trained planner for long-horizon web tasks \cite{arXiv250309572}. Advisor models \cite{arXiv251002453} train a small open-weight advisor with reinforcement learning, on rewards computed from the black-box model's outputs, to write per-instance natural-language advice for that model, and report that an advisor trained with a low-cost student model transfers its improvements to a frontier model. Plan-and-Solve prompting asks a single model to write a plan before solving \cite{arXiv230504091}, and DIN-SQL decomposes text-to-SQL into schema linking, query classification and decomposition, SQL generation and self-correction steps \cite{arXiv230411015}. Several of our mined arms resemble such fixed decompositions, for example BIRD's schema-grounding and CTE-decomposition arms. Weak-to-strong generalization studies the related question of whether a weaker supervisor can elicit a stronger model's capabilities \cite{arXiv231209390}. HSAD keeps the split between a small trained advisor and a frozen executor. It differs from these methods in three ways. First, the advisor is trained by supervised fine-tuning on hindsight targets. Second, the advice is a structured object whose fields take values from closed vocabularies, rather than free text. Third, those vocabularies, and the labels for which strategy helped, come from error analysis of the executor's own failures and task characteristics and from running the executor under different strategies utilizing the training data.

\paragraph{Prompt and context optimization} Automatic prompt engineering searches over instructions with a language model as the proposer \cite{arXiv221101910,arXiv230903409}, and DSPy compiles declarative pipelines by optimizing their prompts and demonstrations \cite{arXiv231003714}. TextGrad treats textual feedback as a gradient \cite{arXiv240607496}, and GEPA evolves prompts by reflecting on rollouts \cite{arXiv250719457}. Context-engineering methods keep an evolving playbook or memory across tasks, for example ACE's generation--reflection--curation loop \cite{arXiv251004618}, Dynamic Cheatsheet's persistent test-time memory \cite{arXiv250407952}, and experience or workflow memories mined from past trajectories \cite{arXiv230810144,arXiv240907429,arXiv250806433}. Self-Refine has the model critique and revise its own output \cite{arXiv230317651}, and Reflexion has an agent keep verbal reflections on failed trials in memory for later attempts \cite{arXiv230311366}. Early experience trains an agent on the consequences of its own exploratory actions, without a reward signal, and reports gains over imitation learning across eight environments \cite{arXiv251008558}. These methods optimize a context that is shared across inputs or accumulated over time. Our arms and directive taxonomy are also a mined, shared vocabulary, but the choice among them, and the values filled into them, are made per input by a trained small model.

\paragraph{Routing and strategy selection} Adaptive-RAG trains a small classifier to predict a question's complexity and routes it to no retrieval, single-step retrieval, or multi-step retrieval \cite{arXiv240314403}. RouteLLM and FrugalGPT route queries between models of different cost \cite{arXiv240618665,arXiv230505176}. The \emph{arm} field of our advice is a router in this sense, and its labels come from running every arm, as in these works. Routing alone, however, leaves gains on the table. A routing decision helps only as much as the predicted best arm is right, and predicting it from the input alone is itself a hard problem. Even with good routing, our results show further gains from the other fields of the advice, which tell the executor what to cover and where to look.

\paragraph{Hindsight relabeling} Hindsight experience replay relabels a failed trajectory with the goal it actually achieved, which turns sparse-reward failures into useful training data \cite{arXiv170701495}. Hindsight instruction relabeling brings the idea to instruction following by relabeling outputs with the instruction they satisfy \cite{arXiv230205206}. For language agents, recent work relabels agent trajectories in hindsight to create successful demonstrations \cite{arXiv260704235}, and uses hindsight over explored experience to create training tasks for web agents \cite{arXiv260627330}. Our labels are hindsight labels of a different kind. We keep the task fixed and relabel which strategy, and which advice, would have made the frozen executor succeed.

\paragraph{Privileged information and context distillation} In learning with privileged information, a teacher sees extra information during training, and a student learns to do well without it. This setting is closely related to knowledge distillation \cite{arXiv151103643}. Context distillation trains a model to give, without a prompt, the answers it gave with the prompt \cite{arXiv211200861,arXiv220915189}. STaR shows a model the correct answer, has it write a rationale for that answer, and trains on the rationale \cite{arXiv220314465}. The idea has long been used to train agents. In Learning by Cheating, a driving agent that sees the ground-truth state of the scene teaches a student that sees only camera images \cite{arXiv191212294}. Several recent methods apply it to language-model agents. $\pi$-Distill trains a teacher that sees privileged information and a student that does not within one model, and uses it to distill frontier agents in multi-turn environments \cite{arXiv260204942}. In LEAP, the teachers can see the environment's hidden state \cite{arXiv241005434}. In gated hindsight distillation, the teacher of a mobile GUI agent also sees the next screenshot, and distillation is applied only when the student fails and the teacher recovers the correct action \cite{arXiv260806065}. Showing the teacher the answer does not always help the student. A teacher that sees a reference answer can pull its targets toward that reference \cite{arXiv260804794}, and in controlled comparisons the correct reference does not help consistently \cite{arXiv260818271}. On-policy self-distillation trains one model as both a teacher that sees privileged information and a student that does not, on the student's own outputs \cite{arXiv260118734}. A study of the method finds that it works when the privileged information is a rule shared across inputs, such as a system prompt, and fails when that information is specific to one input, such as the answer to one problem \cite{arXiv260511182}.

HSAD uses the gold answer and the executor's traces as privileged information. It differs from the methods above in two ways. The student is a separate small model rather than the agent, and it learns to write advice rather than to execute. The difference between shared rules and input-specific facts also matches what we observe. What the advisor learns from the traces, such as which directive fixes which kind of failure, is a rule shared across inputs, and the advisor learns it well. An input-specific fact helps only when the advisor can read it from the input. We therefore build the input-specific fields of the advice around information that the input states, such as the options listed in a shopping instruction. To our knowledge, HSAD is the first method to distill hindsight advice, built from a frozen executor's own failures, into a separate small advisor that writes structured, per-input advice.

\section{Methods}\label{sec:method}

We first define the setting and then describe the five stages of HSAD.

\subsection{Problem setup}\label{sec:setup}

We consider a frozen \emph{executor} $E$ that maps a prompt to an output, and a task distribution over inputs $x$. The executor may be a single model call, as in text-to-SQL, or a multi-step agent that calls tools, as in WebShop and the finance agent. Its base prompt $p_0(x)$ is fixed by the deployment. A scoring function $R(x, o) \in [0,1]$ grades an output $o$ against the gold answer $y^\star(x)$, for example execution accuracy of a SQL query or the fraction of checklist items an answer satisfies. Gold answers exist for a training pool $\mathcal{D}_{\text{train}}$ and for held-out evaluation sets, but not at deployment.

An \emph{advisor} $A_\phi$ is a small language model that reads $x$ and emits advice $z = A_\phi(x)$. The advice is a JSON object whose fields and allowed values are fixed in advance (\S\ref{sec:fields}). A deterministic renderer $r$ turns $z$ into text, which is appended to the base prompt, and the executor runs on $p_0(x) \oplus r(z)$. We measure the method by its \emph{paired lift} on held-out inputs,
\begin{equation}
\Delta \;=\; \frac{1}{|\mathcal{D}_{\text{test}}|}\sum_{x \in \mathcal{D}_{\text{test}}} \Big[ R\big(x, E(p_0(x) \oplus r(A_\phi(x)))\big) - R\big(x, E(p_0(x))\big) \Big],
\end{equation}
in which the same executor answers the same input with and without advice, so that only the advice differs.

At training time, but not at test time, we have privileged information: the gold answer $y^\star$, the executor's traces under each strategy, and the scores those traces received. The teacher that writes advice targets may use all of it. The student advisor sees only $x$. This is the setting of learning with privileged information \cite{arXiv151103643}. Training the student on a target is useful only if the student can produce that target from $x$ alone, and \S\ref{sec:discussion} discusses which parts of the advice meet this condition in our settings.

\begin{figure}[t]
\figblock{\resizebox{\linewidth}{!}{
\definecolor{figTax}{HTML}{7B5CD6}\definecolor{figTaxBg}{HTML}{EEE9FB}
\begin{tikzpicture}[figfont]
\tikzset{ftax/.style={fbox, draw=figTax, fill=figTaxBg}}
\node[facq, text width=3.3cm, anchor=north west] (fail) at (0,0)
  {\textbf{base-executor failures}\\[1pt]{\figfontsmtxt default arm on the train pool,\\graded against gold}};
\node[ffil, text width=3.3cm, below=3.5mm of fail] (desc)
  {\textbf{describe first}\\[1pt]{\figfontsmtxt one free-text root-cause note\\per failed row, no categories yet}};
\node[ffil, text width=3.3cm, below=3.5mm of desc] (merge)
  {\textbf{mine, then merge}\\[1pt]{\figfontsmtxt cluster notes over thousands of\\rows into a closed vocabulary}};
\draw[farr] (fail) -- (desc);
\draw[farr] (desc) -- (merge);

\node[fcon, text width=2.9cm, below left=4.5mm and -1.5cm of merge] (arms)
  {\textbf{arms}\\[1pt]{\figfontsmtxt 5--8 strategies, each a 2--4\\sentence delta appended\\to the base prompt}};
\node[ftax, text width=2.6cm, below right=4.5mm and -1.5cm of merge] (tax)
  {\textbf{directive taxonomy}\\[1pt]{\figfontsmtxt error areas (L1) $\to$\\instantiable directives (L2)}};
\draw[farr] (merge) -- (arms);
\draw[farr] (merge) -- (tax);

\node[fband, anchor=west] (hd) at (6.7,-0.40) {BIRD (text-to-SQL): 8 areas, 63 directives (samples shown below)};
\tikzset{l1/.style={ftax, text width=4.0cm, align=left, anchor=north west},
         l2/.style={figmono, font=\ttfamily\fontsize{6.3}{7.6}\selectfont, anchor=north west, align=left, inner sep=1pt, text width=5.9cm}}
\node[l1] (a8) at (6.7,-3.35) {\textbf{8\ \ flawed evidence over-reliance}};
\node[l2] at (11.35,-3.30) {8.1 strictly\_follow\_evidence\_calculation\_\ldots\\8.3 copy\_evidence\_literals\_verbatim};
\node[l1] (a2) at (6.7,-0.80) {\textbf{2\ \ schema linking and join path}};
\node[l2] at (11.35,-0.75) {2.1 verify\_join\_keys\_against\_foreign\_key\_\ldots\\2.4 join\_descriptive\_tables\_for\_named\_\ldots};
\node[l1] (a7) at (6.7,-2.50) {\textbf{7\ \ dialect and type casting}};
\node[l2] at (11.35,-2.45) {7.1 cast\_operands\_to\_real\_before\_division\\7.3 use\_dialect\_compatible\_date\_functions};
\node[l1] (a5) at (6.7,-1.65) {\textbf{5\ \ projection and cardinality}};
\node[l2] at (11.35,-1.60) {5.1 strictly\_project\_requested\_attributes\\5.4 apply\_distinct\_to\_prevent\_join\_\ldots};

\coordinate (gre) at (17.35,-3.80);
\node[fgroup, fit=(a8)(a5)(hd)(gre), inner sep=4pt] (grp) {};
\draw[farrd] (tax.east) -- ++(0.35,0) |- node[flab, pos=0.75]{e.g.} (grp.west);

\node[fcon, text width=10.3cm, align=left, anchor=north west] (armex) at (6.7,-4.60)
  {\textbf{arm} \texttt{schema\_grounding\_and\_minimal\_joins}: {\figfontsmtxt ``First silently trace the schema graph to find the exact host
   table for every column you will SELECT and every column you will filter on. \ldots\ Before adding any join, check whether the
   column is already available on a table you are using.''}};
\draw[farrd] (arms.south) -- ++(0,-0.2) -| (armex.south);
\end{tikzpicture}}}
\figcap{2}{Error analysis on the base executor's training failures produces two closed vocabularies. Failures are first described in free text and only then clustered. Arms are short strategy deltas appended to the base prompt. Directives are organised as error areas (L1) with specific fixes (L2). Right: four of the eight areas of BIRD's taxonomy, with two sample directives each, and one of its arms, quoted from the arm definition.}
\end{figure}

\subsection{The five stages of HSAD}\label{sec:stages}

HSAD has five stages. The first four run once, offline, over the full training pool: error analysis, hindsight labels, advice targets and advisor training. The fifth, rendering, runs at test time.

\subsubsection{Error analysis and closed vocabularies}\label{sec:mining}

The first stage studies how the executor fails, so that the advice can later be written in terms of those failures. We run the executor with its base prompt over the training pool and grade each output against gold. For each failed input, a strong model writes a short free-text note on what went wrong, without being given any categories. The model sees the input, the executor's output and its grade, together with whatever else explains the failure. For text-to-SQL, for example, it also sees the database schema and the execution error, since an error such as ``no such column'' often points straight to the mistake. We then cluster the notes from thousands of failed inputs into a closed vocabulary. Writing the notes before choosing any categories improves the quality of the vocabulary, since every category comes from failures the executor made on training inputs and every directive addresses one of them.

Two vocabularies come out of this stage (Fig.~2). \emph{Arms} are three to eight whole-response strategies, each a short delta of a few sentences, plus a \cc{default} arm that adds nothing. An arm must add to the base prompt rather than replace it. Otherwise an arm-versus-arm comparison mixes the strategy with incidental rewording of the base instruction. \emph{Directives} form a two-level taxonomy with error areas at the first level and specific, named fixes at the second, for example area 7 of BIRD, ``dialect and type casting'', with directive 7.1 \cc{cast\_operands\_to\_real\_before\_division} in the text-to-SQL taxonomy of Fig.~2. The size of each vocabulary depends on the task (Table~\ref{tab:setup}).

\subsubsection{Hindsight labels from running every arm}\label{sec:labels}

The second stage produces the hindsight label. We run the frozen executor under every arm on every training input, grade every output against gold, and record the winning arm $a^\star(x)$ together with its trace. Some rows produce ties, because several arms often give the same score, and the tie-break is part of the label definition. We break ties toward \cc{default}. If no arm beats doing nothing, the correct advice is to change nothing, and training the advisor to fire an arm on such rows teaches it to intervene without benefit. When scores have several dimensions, ties can be broken by a fixed fallback order of secondary criteria. If every arm fails, or several arms tie and the scores cannot separate them, the teacher's LLM, which sees the gold answer and every arm's output, can pick the arm whose output is closest to correct. Section~\ref{sec:experiments} gives the scheme used in each setting, and Appendix~\ref{app:common} gives examples of how the tie-break changes the advisor.

\subsubsection{Advice targets and the field-source rule}\label{sec:fields}

The third stage writes the advice target $z^\star(x)$ for every training input. This is the signal that the advisor will learn to produce. The \emph{teacher} is the labelling procedure that has hindsight access to gold and to the runs. It combines an LLM, for fields that need judgment, with fixed rules, for fields whose value can be read off the arm scores, the winner's trace or the input. The teacher's LLM receives the input and its grading material (the gold answer or a grading rubric), every arm's output with its score and tool usage, and the full trace of the winning arm. It also receives the closed vocabularies it must choose from: the arms, the directive taxonomy and, where relevant, the values of other fields (e.g., the list of data sources). It then fills a domain-specific JSON schema whose fields are illustrated in Fig.~3 and listed in Table~\ref{tab:fields}. The schema is designed before any LLM call, and the LLM does not fill every field. The quality of this schema matters a great deal for the results. For a new task, iterations on failure analysis and on alternative schemas therefore pay off.

\begin{figure}[t]
\figblock{\resizebox{\linewidth}{!}{
\begin{tikzpicture}[figfont]
\tikzset{
  fld/.style  = {figmono, anchor=north west, inner sep=1.2pt, align=left},
  tagM/.style = {figfontsm, draw=figGood, fill=figGoodBg, rounded corners=1pt, inner sep=1.2pt, anchor=north east},
  tagT/.style = {figfontsm, draw=figFil, fill=figFilBg, rounded corners=1pt, inner sep=1.2pt, anchor=north east},
  tagN/.style = {figfontsm, draw=figMute, fill=figSurf, rounded corners=1pt, inner sep=1.2pt, anchor=north east},
  panel/.style= {draw=figAcq, fill=figSurf2, rounded corners=2pt, line width=0.5pt,
                 minimum width=4.05cm, anchor=north west},
}
\newcommand{\fieldrow}[4]{
  \node[fld, text width=3.9cm] at (0.08,#1) {#2};}

\begin{scope}[shift={(0,0)}]
\node[panel, minimum height=5.85cm] at (0,0.3) {};
\node[figfontlg, anchor=west] at (0.08,0.08) {WebShop};
\node[figfontsm, anchor=north west, text width=3.85cm, align=left] at (0.08,-0.12)
   {``\ldots slim fit men's pants \ldots\ with color: green, and size: xx-large, and price lower than 50.00 dollars''};
\fieldrow{-1.15}{arm: concise\_search\_\\\hspace*{0.8em}and\_fast\_commit}{tagM}{from runs}
\fieldrow{-1.75}{effort: long}{tagM}{from runs}
\fieldrow{-2.15}{expected\_tools:\\\hspace*{0.8em}back\_to\_search,\\\hspace*{0.8em}page\_nav}{tagM}{from runs}
\fieldrow{-3.05}{required\_options:\\\hspace*{0.8em}color: green\\\hspace*{0.8em}size: xx-large}{tagM}{from input}
\fieldrow{-3.95}{directive\_picks:\\\hspace*{0.8em}3.1 select\_color\\\hspace*{0.8em}3.2 select\_size}{tagM}{from input}
\end{scope}

\begin{scope}[shift={(4.3,0)}]
\node[panel, minimum height=5.85cm] at (0,0.3) {};
\node[figfontlg, anchor=west] at (0.08,0.08) {BIRD};
\node[figfontsm, anchor=north west, text width=3.85cm, align=left] at (0.08,-0.12)
   {``What is the average score of the posts owned by the user csgillespie?''};
\fieldrow{-1.15}{arm: schema\_grounding\_\\\hspace*{0.8em}and\_minimal\_joins}{tagT}{teacher}
\fieldrow{-1.75}{effort: high}{tagM}{from runs}
\fieldrow{-2.15}{length: medium}{tagM}{from runs}
\fieldrow{-2.55}{directive\_picks:\\\hspace*{0.5em}1.1 validate\_attribute\_\\\hspace*{1.6em}existence\_and\_ownership\\\hspace*{0.5em}2.4 join\_descriptive\_tables\_\\\hspace*{1.6em}for\_named\_entities\\\hspace*{0.5em}8.1 strictly\_follow\_evidence\_\\\hspace*{1.6em}calculation\_formulas\\\hspace*{0.5em}8.3 copy\_evidence\_literals\_\\\hspace*{1.6em}verbatim}{tagT}{teacher}
\end{scope}

\begin{scope}[shift={(8.6,0)}]
\node[panel, minimum height=5.85cm] at (0,0.3) {};
\node[figfontlg, anchor=west] at (0.08,0.08) {ExpertLongBench};
\node[figfontsm, anchor=north west, text width=3.85cm, align=left] at (0.08,-0.12)
   {an agent trajectory to audit for security threats (task T11CyberRDG)};
\fieldrow{-1.15}{arm: security\_threat\_\\\hspace*{0.8em}frame\_first}{tagM}{from runs}
\fieldrow{-1.75}{target\_length: short\\\hspace*{0.8em}(<300 words)}{tagM}{from runs}
\fieldrow{-2.35}{effort: low}{tagM}{from runs}
\fieldrow{-2.75}{directive\_picks:\\\hspace*{0.5em}1.5 analyze\_security\_\\\hspace*{1.6em}trajectory\_not\_surface\_\ldots\\\hspace*{0.5em}2.5 detect\_adversarial\_\\\hspace*{1.6em}content\_in\_tool\_returns\\\hspace*{0.5em}3.10 populate\_all\_security\_\\\hspace*{1.6em}analysis\_fields\\\hspace*{0.5em}8.1 detect\_and\_label\_\\\hspace*{1.6em}prompt\_injection}{tagT}{teacher}
\end{scope}

\begin{scope}[shift={(12.9,0)}]
\node[panel, minimum height=5.85cm] at (0,0.3) {};
\node[figfontlg, anchor=west] at (0.08,0.08) {FrontierFinance};
\node[figfontsm, anchor=north west, text width=3.85cm, align=left] at (0.08,-0.12)
   {Colgate-Palmolive revenue by geography and product, each quarter since 2019};
\fieldrow{-1.15}{arm: default}{tagM}{from runs}
\fieldrow{-1.55}{target\_length: long\\\hspace*{0.8em}(3+ pages)}{tagT}{teacher}
\fieldrow{-2.15}{web\_use: avoid}{tagT}{teacher}
\fieldrow{-2.55}{sources: FILINGS, \ldots}{tagT}{teacher}
\fieldrow{-2.95}{tools: documents\_list,\\\hspace*{0.8em}documents\_search, \ldots}{tagT}{teacher}
\fieldrow{-3.55}{directive\_picks:\\\hspace*{0.5em}2.2 distinguish\_metric\_\\\hspace*{1.6em}unavailability\_from\_zero\\\hspace*{0.5em}1.1 enumerate\_all\_periods\_\\\hspace*{1.6em}in\_requested\_range\\\hspace*{0.5em}\ldots}{tagT}{teacher}
\end{scope}

\end{tikzpicture}}}
\figcap{3}{Example advice JSON in the four domains. Table~\ref{tab:fields} lists where the training label of each field comes from. Directive identifiers refer to the domain's taxonomy (Fig.~2).}
\end{figure}

\paragraph{Where each label comes from} Every training label has two properties: how the teacher computes it and what it is computed from. It is computed either by one of the teacher's fixed rules (\emph{rule-based}) or by the teacher's LLM (\emph{LLM reasoning}). It is computed from the input alone, from the gold answer, or from the executor's runs. From the runs, a rule can read either the arm scores, meaning the grader's score of every arm's output against gold, or the winner's trace, meaning the run of the winning arm. Everything except the input is available only in hindsight. Table~\ref{tab:fields} gives both properties for every field we use.

\paragraph{Rules where possible, the LLM where necessary} The same field can often be obtained in either way. Answer length, for example, can be computed by a rule from the winner's trace or estimated by the teacher's LLM. We use a rule whenever a field's value can be read off the runs, the gold answer or the input, and we use the teacher's LLM only for fields that require judgment. We adopted this rule after finding that the LLM's estimates can disagree with the runs (\S\ref{sec:elb}). Effort and length fields are therefore coarse buckets of the winning trace's tool-call count, output complexity or answer length. The arm is always the hindsight winner, and the LLM picks it only when the scores cannot separate the arms.

\paragraph{Training labels and test time} The source of a training label is separate from how the field is produced at test time. A label computed by a rule on the runs, such as the highest-scoring arm, still has to be predicted by the advisor, because the runs it was computed from exist only after the executor has run. In all four settings the trained advisor predicts every field from the input alone (last column of Table~\ref{tab:fields}). Only a rule on the input could in principle replace the advisor for a field, as for the option fields on WebShop (\S\ref{sec:webshop}).

\paragraph{Leakage} The teacher's LLM sees the gold answer, so the advice it writes must not leak that answer, for example by naming the value the answer should contain. We prevent this in how the vocabularies are built and used. When we mine a directive taxonomy, the prompt requires every directive to be a generic behaviour that applies across many inputs, with no entity, date, period or value in it, and candidates that name one are rewritten or dropped. When the teacher's LLM writes an advice target, it only picks among these fixed values, and the executor receives the fixed description of each picked directive. The advice therefore contains no free text written with the gold answer in view.

\begin{table}[t]
\centering\small
\resizebox{\linewidth}{!}{\begin{tabular}{@{}llllll@{}}
\toprule
& & \multicolumn{3}{c}{Training label} & At test time \\
\cmidrule(lr){3-5}
Domain & Field & computed by & computed from & how & produced by \\
\midrule
WebShop & \cc{arm} & rule-based & arm scores & highest-scoring arm; ties to \cc{default} & advisor \\
 & \cc{effort}, \cc{expected\_tools} & rule-based & winner's trace & winner's tool calls & advisor \\
 & \cc{required\_options}, \cc{directive\_picks} & rule-based & input & options stated in the instruction & advisor \\
\midrule
FrontierFinance & \cc{arm} & rule-based & arm scores & highest-scoring arm; cascade tie-break & advisor \\
 & \cc{target\_length}, \cc{web\_use}, \cc{sources}, \cc{tools} & LLM reasoning & gold + all runs & reads all arms' answers, grades & advisor \\
 & \cc{directive\_picks} & LLM reasoning & gold + all runs & 103 directives & advisor \\
\midrule
BIRD & \cc{arm} & LLM reasoning & gold + all arms' queries, results & arm closest to gold & advisor \\
 & \cc{effort}, \cc{length} & rule-based & winner's trace & winning SQL's structure, length & advisor \\
 & \cc{directive\_picks} & LLM reasoning & gold + all runs & 63 directives & advisor \\
\midrule
ExpertLongBench & \cc{arm} & rule-based & arm scores & highest-scoring arm; ties to \cc{default} & advisor \\
 & \cc{target\_length}, \cc{effort} & rule-based & winner's trace & winner's length & advisor \\
 & \cc{directive\_picks} & LLM reasoning & gold + all runs & 57 directives & advisor \\
\bottomrule
\end{tabular}}
\caption{Source of each advice field's training label: how the teacher computes it (rule-based or by LLM reasoning) and what it is computed from. ``Arm scores'' are the grader's scores of every arm's output against gold; ``winner's trace'' is the run of the winning arm. The last column shows what produces each field at test time.}
\label{tab:fields}
\end{table}

\subsubsection{Advisor training and selection}\label{sec:sft}

The fourth stage trains the advisor by supervised fine-tuning. Each training example pairs an input with the teacher's advice target for it. The advisor is a small open model that learns to produce the JSON target from the input and a system prompt that lists the schema and vocabularies. It never sees gold. Because it is small, it is cheap to train, and running it adds one low latency model call per input. We select the epoch by paired lift on a held-out validation split, running the executor with each checkpoint's advice, rather than by the agreement between the advisor's JSON and the targets (see Appendix~\ref{app:common} for more details).

\subsubsection{Rendering}\label{sec:render}

The fifth stage turns the advice into text for the executor. Each field has a fixed template. The arm contributes its strategy delta, effort and length become one sentence each (``Plan for 6 or more tool calls in total.''), and each directive contributes its one-sentence description. The rendered text is appended to the executor's prompt or message list, as guidance that the executor is free to follow or ignore.

\paragraph{Instantiated directives} A directive can be generic (``select every required option before buying'') or \emph{instantiated} with values from the input (``Before clicking Buy Now you MUST click the size button labelled `xx-large'\,''). A generic directive asks the executor to work out what to do, whereas an instantiated one is more specific.

\subsection{Performance comparisons}\label{sec:ceilings}

To judge how much of the possible gain the trained advisor captures, we compare it with reference points that use hindsight. Studies on advice and routing often call such a reference an ``oracle'', but the term is used for different quantities, so we define ours here.
\begin{enumerate}[leftmargin=*]
\item The \emph{best-arm oracle} is the highest score that any arm received during collection, averaged over the dataset. Like the other two reference points, we report it as a lift over the \cc{default} arm. It bounds the gain from choosing the arm, but not the gain from the full advice.
\item The \emph{teacher-executed ceiling} runs the executor again on held-out inputs with the teacher's hindsight advice. It leaves out the advisor's errors in predicting this advice target.
\item The \emph{student} is the trained advisor's advice, predicted from the input alone and executed. It measures the deployable performance, as the lift over the executor without advice.
\end{enumerate}

\section{Experimental Setup}\label{sec:experiments}

We evaluate HSAD in four settings that differ in the executor, the kind of output and the scoring function. Our main setting is FrontierFinance, where the executor is a production research agent and the goal is to improve on a strong proprietary system. The other three settings use a small open executor and test whether the method carries over to shopping, text-to-SQL and expert long-form generation. Table~\ref{tab:setup} summarizes them, and Appendix~\ref{app:common} gives further details of the experimental setup.

\begin{table}[t]
\centering\small
\resizebox{\linewidth}{!}{\begin{tabular}{@{}lllrrrr@{}}
\toprule
Setting & Frozen executor & Metric & Train & Test & Arms & Directives \\
\midrule
FrontierFinance & Samaya production agent & rubric qualification rate & 1,819 & 220 & 4 & 103 \\
WebShop & Qwen3-4B agent & score (0--100), success rate & 6,210 & 500 & 6 & 6 \\
BIRD & Qwen3-4B & execution accuracy & 9,421 & 1,074 & 9 & 63 \\
ExpertLongBench & Qwen3-4B & checklist coverage & 450 & 100 & 8 & 57 \\
\bottomrule
\end{tabular}}
\caption{The four settings. ``Train'' is the number of advisor training rows. Arms include \cc{default}.}
\label{tab:setup}
\end{table}

\paragraph{FrontierFinance} FrontierFinance is a public benchmark of 220 expert-written financial research queries, graded against 11,543 source-attributed rubric items that span six use cases of an investor's workflow, from earnings and events to screening and discovery \cite{arXiv260811683}. Each query has its own rubric, and some rubric items are marked essential. The metric is the qualification rate, the fraction of rubric items an answer satisfies, averaged over queries. We report it for all items, as the public leaderboard does, and separately for essential items. Our executor is Samaya's production research agent, which leads the benchmark's leaderboard \cite{arXiv260811683}, with tools over financial filings, market-data services and the web. We freeze it and try to improve it only through advice. Its four arms are \cc{default}, \cc{structured\_priority} (prefer structured and tabular data tools), \cc{tables\_extraction} (extract figures from filing tables), and \cc{restricted\_tools} (structured sources only, no web search). Ties between arms are broken by a fallback order: essential criteria, then all criteria, then cost, then latency. The advisor is trained on 1,819 questions from the same expert-annotated pool that FrontierFinance was drawn from, none of which is among the 220 public queries. It is evaluated on a separate held-out validation set and on the 220 public queries, which serve as the test set.

\paragraph{WebShop} WebShop is a simulated e-commerce website in which an agent receives a shopping instruction, such as this one from its test set: ``slim fit men's pants with elastic waist \ldots\ with color: green, and size: xx-large, and price lower than 50.00 dollars'', and must search, open products, select options and buy \cite{arXiv220701206}. The environment's reward gives partial credit for the product type, its attributes, the selected options and the price. We report it as a score from 0 to 100, together with the success rate, the fraction of episodes with a perfect reward. The executor is a frozen Qwen3-4B agent that emits one JSON action per turn, with a 15-turn cap and a fixed environment seed. The training pool has 6,410 goals (samples of the dataset), of which 200 are held out for epoch selection, and the test set has 500 samples.

\paragraph{BIRD} BIRD is a text-to-SQL benchmark of 12,751 question--SQL pairs over 95 databases \cite{arXiv230503111}. Each question comes with an ``evidence'' hint that states domain knowledge, for example that ``Italian refers to language = `Italian'\,''. The metric is execution accuracy, the fraction of questions whose predicted query returns the same result as the gold query. The executor is a frozen Qwen3-4B that writes one query per question. The training pool is BIRD's training split (9,428 questions over 69 databases). We split BIRD's development set into 460 questions for epoch selection and 1,074 for test. Because many training questions have several correct arms or none, the teacher's LLM, which sees the gold query and every arm's query and result, picks the arm closest to correct.

\paragraph{ExpertLongBench} ExpertLongBench is a benchmark of expert-level long-form tasks, each with a rubric designed or validated by domain experts \cite{arXiv250601241}. Outputs are graded by extracting a checklist of the rubric's items from the output and comparing it with the checklist extracted from a reference. We use six of its public tasks: materials-science synthesis extraction (T03), pedagogical evaluation of tutoring dialogues (T04), clinical note generation (T06), molecule and protein description generation (T07, T08) and security review of agent trajectories (T11). We exclude the seventh public task (legal multi-document summarization), whose source documents have a median length of about 62K tokens. The metric is the fraction of checklist fields an answer satisfies. The executor is a frozen Qwen3-4B. It has 450 training and 100 test documents.

\section{Results}\label{sec:results}

We report results on FrontierFinance, our main setting, and then on WebShop, BIRD and ExpertLongBench. The advice improves the executor in all four settings (Tables~\ref{tab:ff-leaderboard} and~\ref{tab:main}). For each setting, we also examine which advice fields contribute to the gain, and give additional analyses in Appendix~\ref{app:details}.

\subsection{FrontierFinance}\label{sec:qa}

\begin{table}[t]
\centering\small
\begin{tabular}{@{}lrrr@{}}
\toprule
System (harness) & Qualification rate & Cost / query & Latency / query \\
\midrule
Samaya System + advisor & \textbf{56.6\%} & \$1.14 & 257 s \\
Samaya System, high effort & 56.0\% & \$1.81 & 278 s \\
Claude Fable 5.1 (Finance Agent v2) & 55.9\% & \$6.89 & 268 s \\
Samaya System (no advice) & 52.9\% & \$0.93 & 214 s \\
Claude Fable 5 (Finance Agent v2) & 49.2\% & \$4.06 & 165 s \\
GPT 5.6 Sol (Finance Agent v2) & 46.8\% & \$3.03 & 171 s \\
Kimi K3 (Finance Agent v2) & 46.4\% & \$0.90 & 336 s \\
Gemini 3.6 Flash (Finance Agent v2) & 46.3\% & \$2.41 & 164 s \\
Claude Opus 4.8 (Finance Agent v2) & 45.0\% & \$2.61 & 156 s \\
\bottomrule
\end{tabular}
\caption{FrontierFinance public leaderboard (220 queries, all rubric items) with the advised agent added. All rows except the advised agent are taken from the public FrontierFinance leaderboard \cite{arXiv260811683}. The two Samaya rows from the leaderboard use the same proprietary harness we use, the other rows use the open-source Finance Agent v2 harness.}
\label{tab:ff-leaderboard}
\end{table}

The finance setting asks whether a small advisor can still improve a system that is already strong. The executor leads the benchmark's leaderboard, and its harness was built for this kind of research, so any gain has to come from advice that the agent does not already give itself. Each run of the agent takes about four minutes and costs about a dollar, so the arms collection over 1,819 training questions is the most expensive step of the pipeline.

The advice has six fields (Fig.~3). The \cc{arm} is the hindsight winner strategy. The teacher's LLM writes \cc{target\_length} (short, medium or long, with fixed page counts for each category), \cc{web\_use} (\cc{avoid}, \cc{fallback} or \cc{required}), a prioritized list of data \cc{sources}, a list of suggested \cc{tools}, and \cc{directive\_picks} drawn from a 103-directive taxonomy in 9 areas (for example 3.1 \cc{cover\_all\_named\_entities\_in\_query}). To write these fields, the LLM sees each training question's rubric, the four arms' answers with their rubric scores, tool calls and costs, and the tool-call trace of the winning arm. The rendered advice is appended to the query as a guidance block (Appendix~\ref{app:render}). It is phrased as a suggestion, with an explicit fallback to the agent's own exploration when needed.

\label{sec:qa-leaderboard}Table~\ref{tab:ff-leaderboard} compares the performance of the advised agent against the leaderboard's entries. The leaderboard's Samaya System entry is the same agent without advice. With the advisor, the agent's qualification rate rises from 52.9\% to 56.6\%. This is above the Samaya System's high-effort configuration (56.0\%), which adds improved tools, stronger context management and more reasoning effort, and above the best frontier model in the open finance-agent harness (Claude Fable 5.1, 55.9\%). The advised agent costs \$1.14 per query, against \$0.93 without advice, \$1.81 for the high-effort configuration and \$6.89 for Claude Fable 5.1. Fig.~4 shows the same comparison as a cost--quality trade-off. The advisor moves the Samaya System up to the top of the leaderboard while staying in the low-cost region, where every other system of similar cost, apart from the unadvised agent itself, scores at least ten points lower. More details are included in Appendix~\ref{app:ff-deploy}.

\label{sec:qa-ablation}To find out which fields carry the gain, we ran a leave-one-field-out ablation on the validation set. The ablation uses the teacher's own advice rather than the student's, so that each run isolates one field without prediction errors. Table~\ref{tab:ff-ablation} reports the results. With all fields, the teacher's advice scores 69.3\% essential against 63.0\% without advice, a teacher-executed gain of 6.3pp that the trained advisor matches (6.2pp). Dropping the directive picks removes the whole gain. Every other field contributes a smaller amount, between 1.8pp and 4.4pp. The ablation was run while the schema still had an \cc{effort} field. Removing it slightly improved the score, so the final schema does not include it.

\begin{table}[H]
\centering\small
\begin{tabular}{@{}lrrr@{}}
\toprule
Teacher advice & essential & all & Contribution (essential) \\
\midrule
all fields & 69.3\% & 59.6\% & -- \\
\quad without \cc{directive\_picks} & 62.6\% & 54.3\% & 6.7pp \\
\quad without \cc{sources} & 64.9\% & 55.5\% & 4.4pp \\
\quad without \cc{tools} plan & 66.3\% & 56.2\% & 3.0pp \\
\quad without \cc{web\_use} & 66.6\% & 57.0\% & 2.7pp \\
\quad without \cc{target\_length} & 67.5\% & 57.5\% & 1.8pp \\
\quad without \cc{effort} & 70.0\% & 59.0\% & $-0.7$pp \\
no advice & 63.0\% & 53.9\% & -- \\
\bottomrule
\end{tabular}
\caption{Leave-one-field-out ablation of the teacher's advice on the FrontierFinance validation set. The contribution of a field is the drop in essential-criteria coverage when it is removed.}
\label{tab:ff-ablation}
\end{table}

\begin{figure}[t]
\figblock{
\begin{tikzpicture}[figfont]
\begin{axis}[
  width=9.5cm, height=5.0cm,
  xmode=log, xmin=0.2, xmax=9, ymin=29, ymax=62,
  xtick={0.25,0.5,1,2,4,8}, xticklabels={\$0.25,\$0.5,\$1,\$2,\$4,\$8},
  ytick={30,40,50,60}, yticklabels={30\%,40\%,50\%,60\%},
  xticklabel style={figfontsm}, yticklabel style={figfontsm},
  xlabel={\figfontsmtxt cost per query (log scale)},
  ylabel={\figfontsmtxt qualification rate (all rubric items)},
  axis line style={figMute}, tick style={figMute},
  grid=major, major grid style={figSurf},
  clip=false,
]
\addplot[only marks, mark=diamond*, mark size=2.6pt, draw=figInk2, fill=figSurf] coordinates {
 (2.6109,45.04) (4.0592,49.19) (6.89,55.86) (0.9487,32.32) (0.9012,46.37) (0.6276,42.78)
 (0.677,40.54) (2.8005,43.47) (3.0324,46.83) (1.7228,30.51) (2.4091,46.27)};
\addplot[only marks, mark=*, mark size=2pt, draw=figMute, fill=white] coordinates {
 (1.4737,32.98) (0.2406,30.72)};
\addplot[only marks, mark=star, mark size=3.4pt, draw=figAcq, line width=0.8pt] coordinates {
 (0.93,52.87) (1.8093,56.04)};
\addplot[only marks, mark=star, mark size=4.2pt, draw=figGood, fill=figGood, line width=1pt] coordinates {(1.1443,56.56)};
\draw[farrd] (axis cs:0.93,52.87) -- (axis cs:1.10,56.2);
\node[figfontsm, anchor=east, text=figAcq] at (axis cs:0.88,51.3) {Samaya};
\node[figfontsm, anchor=west, text=figAcq] at (axis cs:1.9,55.2) {Samaya, high effort};
\node[figfontsm, anchor=east, text=figGood, font=\sffamily\bfseries\fontsize{6.8}{8}\selectfont] at (axis cs:1.06,58.3) {Samaya + advisor};
\node[figfontsm, anchor=west] at (axis cs:6.95,55.4) {Fable 5.1};
\node[figfontsm, anchor=west] at (axis cs:4.1,49.2) {Fable 5};
\node[figfontsm, anchor=north] at (axis cs:0.9012,45.8) {Kimi K3};
\end{axis}
\end{tikzpicture}}
\figcap{4}{Qualification rate against cost per query on the FrontierFinance public leaderboard (220 queries). Stars are the proprietary Samaya harness, diamonds the open-source Finance Agent v2 harness with different frontier and open-weight models, and circles a web-search-only harness. The green star is the Samaya System with the trained advisor; the dashed arrow connects it to the same system without advice.}
\end{figure}

The directives are general guidance, for example to give every named company equal depth of coverage or to pull sub-segment detail within reported segments. Such guidance is the main source of the gain for this finance task. In Appendix~\ref{app:examples}, we show two test queries with the advisor's output and its effect on the answer. On the training side, further variants, namely more epochs, a reasoning trace before the JSON and RL, did not improve performance, as we discuss in Appendix~\ref{app:ff-labels}. 

Taken together, these results show that an 8B advisor trained on hindsight labels lifts Samaya's production agent from 52.9\% to 56.6\% on the public FrontierFinance leaderboard, above the agent's own high-effort configuration at 37\% lower cost per query. 

\subsection{WebShop}\label{sec:webshop}

\begin{table}[t]
\centering\small
\begin{tabular}{@{}llrrr@{}}
\toprule
Setting & Metric & No advice & With advice & Lift \\
\midrule
WebShop & score / success rate & 54.62 / 3.8\% & 63.35 / 23.2\% & +8.73 / +19.4 \\
BIRD & execution accuracy & 40.04\% & 43.11\% & +3.07 \\
ExpertLongBench & checklist coverage & 49.7\% & 54.5\% & +4.9 \\
\bottomrule
\end{tabular}
\caption{Held-out lift of the trained advisor with a frozen Qwen3-4B executor. Each row compares the same executor on the same inputs with and without advice. Success-rate lifts are in percentage points.}
\label{tab:main}
\end{table}

On WebShop, the trained advisor raises the executor's score from 54.6 to 63.4 and its success rate from 3.8\% to 23.2\% (Table~\ref{tab:main}). We reached this design in several steps (Appendix~\ref{app:ws-designs}). Advisors that only chose the arm gained at most 3.65 points, and adding generic directives initially lowered the score before further optimizations.

To find what the arms were missing, we measured where the lost reward went. Across the 5,927 training samples where even the best arm fell short of full reward, 69\% of the lost reward came from missed options, 18\% from attributes, 10\% from product type and 3\% from search or commitment failures. The executor could search and it could click, but it often did not work out which option buttons the instruction required. A generic directive, ``inspect all option buttons and click every required one'', did not help. An instantiated one did. Parsing the option values from the instruction and rendering them as ``Before clicking Buy Now you MUST click the size button labelled `x-large'; \ldots'' raised score by 8.9 points on validation. The option values are stated verbatim in the instruction, and a regular expression recovers these values for every sample. Rendering the values as click instructions removes the extraction step, and the executor only has to click the named buttons.

The final advisor emits the schema given in Fig.~3. All of its training labels come from the teacher's rules, without any LLM. The arm, effort and expected tools come from the scored runs, and the options and their directive IDs are parsed from the instruction, where they are stated verbatim and match the gold goal. On 200 validation samples the advisor's \cc{required\_options} matched the instruction's option values on every sample. On test, its full advice gains 8.73 score points. The option values could also be filled in by a rule, but the arm, effort and tool fields come from hindsight and must be learned; with them, the success rate rises from 3.8\% to 23.2\%, against 19.4\% with the options alone.

Our test set is WebShop's standard 500 test instructions, the same set used by ReAct \cite{arXiv221003629}, so the scores are comparable, although the agent harnesses differ. With advice, the Qwen3-4B agent's score (63.4) moves above the imitation- and reinforcement-learning agent of the original paper (62.4) \cite{arXiv220701206} and stays below ReAct prompting with a 540B model (66.6) \cite{arXiv221003629}. Its success rate, 23.2\% against ReAct's 40.0\%, is still much lower, so the agent with a small executor more often buys products that match the instruction partly rather than exactly.

\subsection{BIRD}\label{sec:bird}

On BIRD, advice must steer a single SQL generation, and the most useful knowledge is which tables, columns and joins the question needs. This is harder for an advisor than WebShop's options, because it is not stated in the question, and in our setup the BIRD advisor sees only the question and its evidence hint, not the database schema.

The trained advisor raises execution accuracy from 40.04\% to 43.11\% on the 1,074 test questions. Appendix~\ref{app:bird-ref} compares the result against the three reference points. The best arm recorded during collection would add 20.11 points, running each question again with its best arm adds 11.27, and the student captures about 27\% of the executed ceiling. By the advisor's chosen arm, the largest gains are on \cc{schema\_grounding\_and\_minimal\_joins} (+7.98 on 188 questions) and \cc{strict\_projection\_and\_output\_alignment} (+13.16 on 38). One example query is the following: ``What is the average score of the posts owned by the user csgillespie?'' The executor without advice filters on a denormalized display-name column of the posts table and returns the wrong answer. With the advice of Fig.~3 (schema grounding, ``join descriptive tables for named entities'') it joins the users table on the owner's identifier and filters on \cc{users.DisplayName}, which matches the gold result. Another pattern is dialect. Given ``how many badges \ldots\ were obtained in the year 2011'', the executor wrote \cc{YEAR(Date) = 2011}, which SQLite does not support, and the advice ``use dialect-compatible date functions'' led it to \cc{strftime('\%Y', Date) = '2011'}.

We also analysed the advisor's own errors on the test set (Appendix~\ref{app:c-bird}). When the advice hurts, it is usually because the executor applies a directive where it does not fit, for example by adding an unneeded join. When the executor answers a question wrongly both with and without advice, an existing directive already covers the root cause in 94\% of the cases we inspected. The remaining gap therefore lies mostly in choosing and applying directives, and few failures point to a missing strategy. The advised executor can be compared with published work, although published BIRD results use the full development set or the hidden test set, while ours use a 1,074-question subset of the development set, so comparisons are approximate. The advised Qwen3-4B executor performs comparably to what BIRD's authors report for ChatGPT on the test set with external knowledge (39.3\%, or 40.1\% with chain-of-thought prompting) \cite{arXiv230503111}, and below DIN-SQL (55.9\% on the test set) \cite{arXiv230411015}, which prompts GPT-4 to solve the task in separate steps (schema linking, query classification and SQL generation) and then to correct its own query.

The advice thus helps on BIRD, mainly by grounding questions in the schema (Appendix~\ref{app:c-bird}), but the student captures only about a quarter of the executed ceiling. These analyses show that there is room for improvement with further iterations.

\subsection{ExpertLongBench}\label{sec:elb}

ExpertLongBench tests the method where data is scarce and outputs are long. The advisor has only 450 training documents across six tasks, and each output is a structured report of several hundred words.

Error analysis produced eight arms, several of which are task-specific frames such as \cc{security\_threat\_frame\_first}, \cc{pedagogical\_evaluator\_role\_lock} and \cc{clinical\_note\_completeness\_and\_fidelity}, and a 57-directive taxonomy in nine areas. The areas cover task interpretation, source grounding, field completeness, value accuracy, schema-field disambiguation, negative findings, output structure, security analysis and pedagogical evaluation. Example directives are 2.2 \cc{scan\_entire\_source\_for\_each\_field}, 6.1 \cc{record\_absent\_fields\_explicitly} and 8.1 \cc{detect\_and\_label\_prompt\_injection}. The best-arm oracle over the eight arms is +9.5 points on the training pool. The \cc{target\_length} labels come from the winning answer's length. We first let the teacher's LLM estimate this field, but its estimate matched the length of the winning answer on only 149 of 450 training rows (33\%), and it systematically called for longer answers than the winning arms produced. Replacing the estimates with the winner's actual length bucket changed 301 of the 450 labels.

Trained on the vocabularies and labels described above, the advisor raises checklist coverage from 49.7\% to 54.5\% on the 100 test documents. We analyse the distribution of the lift over tasks, together with the advisor's arm and directive choices for each task, in Appendix~\ref{app:elb-tasks}. In short, the advisor picks directives that fit each task but sometimes chooses an arm meant for a different task. With only about 75 training documents per task, this is not surprising. We expect more training data, and an error analysis done separately for each task, will improve the arm choices and raise the gain further. 

\section{Discussion}\label{sec:discussion}

\paragraph{Types of information in advice} The advice carries two kinds of information. The first is insight learned from the executor's traces. Error analysis on the training pool shows how this executor fails and which strategies fix each failure, and the hindsight labels show which fix actually worked on which input. The advisor turns this into a per-input choice from the arms and directives. This kind of information is not in the input, and the advisor could not have produced it without seeing the traces during training. All three settings that use mined directives show this. On FrontierFinance, the directives distilled from the agent's failed and successful traces are what the gain depends on: without them, the teacher's advice no longer improves on no advice (Table~\ref{tab:ff-ablation}). On ExpertLongBench, the same kind of directives lets the advisor pick guidance that fits each task, and on BIRD, a mined directive already addresses the root cause of 94\% of the remaining failures.

The second kind is instance-specific values filled into the advice. WebShop's option values are stated in the instruction, so the advisor copies this field without error, while the arm, effort and tool fields are what it learns from hindsight. On BIRD, the most useful values are the tables and columns a query needs. Our advisor does not see the database schema, so it has to infer them from the question, which is the same inference the executor gets wrong, and the student captures only about a quarter of the executed gain. This matches two observations from prior work on privileged information. A student can only learn what a teacher infers from privileged information if it can recover the same thing from its own input \cite{arXiv241005434}, and on-policy self-distillation works when the privileged information is a shared rule, such as a system prompt, but fails in the settings tested when it is specific to each instance and absent at test time \cite{arXiv260511182}. In our setting, the insight from the traces plays the role of shared rules, and the filled-in values are instance-specific facts, which the advisor can supply only when the input contains them.

\paragraph{Specific instructions work better than guidance for small executors} In both Qwen3-4B settings, advice that tells the executor exactly what to do helped more than advice that asks it to reason more carefully. On WebShop, ``click the size button labelled `xx-large'\,'' gained almost 9 points, while ``inspect all option buttons'' did not help and ``verify the product type'' hurt. On BIRD, most of the harm came from directives that the executor applied where they did not fit. One reading is that a small executor does not have spare capacity to decide when a general piece of guidance applies, so general guidance mostly adds text to its prompt, whereas a specific instruction replaces a decision it would otherwise get wrong. The finance agent, whose executor is a large frontier model, is the exception. Its gain comes mostly from general directives such as ``cover all named entities in the query'' (\S\ref{sec:qa-ablation}), which is consistent with the same reading, since a stronger executor can decide what to do with such guidance.

\paragraph{Label quality matters} In our experiments, the way the training labels were defined had a clear effect on the trained advisor. On FrontierFinance, the cascade rule for the winning arm changed 681 of the 1,819 labels and gave our best advisor. More epochs, a reasoning trace and RL did not improve on it (Appendix~\ref{app:c-ff}). On WebShop, breaking ties toward \cc{default} gave a better advisor than breaking them toward the arm with the fewest turns (Appendix~\ref{app:common}). On ExpertLongBench, the teacher's LLM estimated the answer length correctly for only a third of the training samples. An advisor trained on these estimates would have learned mostly the LLM's bias. All three cases follow the field-source rule of \S\ref{sec:fields}. Some fields have a single correct value that a rule can compute, such as the answer length or the options stated in the instruction. For these fields, the rule gives more reliable labels than the LLM. Other fields, such as the directives, require judgment across several runs. For these, the LLM's labels carry the gain, as the FrontierFinance ablation shows.

\paragraph{Advice quality depends on the data and the task} The same pipeline gave gains of different size in the four settings. The differences come mainly from the training data and the task. The advisor can only be as good as its labels and as broad as its training pool. On ExpertLongBench, for example, 450 training documents were enough to learn task-appropriate directives but not enough to learn a reliable arm choice. The advice also helps only where the executor's traces show failures that advice can fix. An instantiated field helps only if its value can be recovered from the input. WebShop's option values can be recovered this way, but BIRD's tables and columns cannot. Before applying the method to a new setting, we suggest checking four things. The executor should succeed on some training inputs and fail on others. The training pool should be large and reliably graded. Error analysis should reveal failures that advice can address. Finally, any instance-specific values that the advice needs should be recoverable from the input.

\paragraph{Advice as guidance} The advice is appended to the executor's instructions as guidance. The executor keeps all of its tools and can depart from the advice when the evidence points elsewhere. Not every suggestion is followed, but on average the advice helps. The fields also differ in how closely they are tied to one executor. Some fields describe the input and what a good answer needs, such as which sources hold the answer, how many periods it spans or which disclosures to check. These fields should carry over to other executors. Other fields reflect the habits of the executor whose traces the advisor learned from, such as its typical blind spots. These fields are most useful for that executor.

\paragraph{Advice in the prompt rather than in the weights} Our advice is text, and the advisor acts as a small adapter that supplies task know-how for each input. A single optimized prompt gives every input the same guidance. The advisor instead gives each input its own guidance, without making every prompt longer. The executor never changes, so the method works with closed or shared models, and the same advice can be given to a different executor. The cost is that the executor reads the advice on every call and does the extra work it asks for. On the FrontierFinance leaderboard, the advised agent costs 23\% more per query than the agent without advice, but it still costs 37\% less than the agent's high-effort configuration and scores higher, which shows the value of guidance tailored to each input. Methods that distil experience or skills into the executor's weights \cite{arXiv260721051,arXiv260809826}, or that have a side model produce parameters or latent tokens instead of text \cite{arXiv250606105,arXiv260215902,arXiv250924704}, avoid this per-call cost, but they have to modify the executor. Advice written as text also has an interpretability benefit, since the advice given for each input can be read and inspected directly.

\section{Limitations}\label{sec:limitations}

\paragraph{Statistical power} Each result comes from a single evaluation run per condition. Most of our test sets are large enough that the averages over samples are reasonably stable: 220 queries on FrontierFinance, 500 instructions on WebShop and 1,074 questions on BIRD. The exception is ExpertLongBench, with 100 test documents. Our validation sets are also small, with 100 questions on FrontierFinance and 200 samples on WebShop, so differences of a few points between advisors on validation can be within run-to-run variation. Repeated runs with different seeds would give confidence intervals for every comparison, and we leave this to future work.

\paragraph{Iterations} The arms and the directive taxonomies went through several rounds of revision, each guided by error analysis of the executor's traces, and most rounds improved the result. On WebShop, rebuilding the taxonomy around instantiated option directives raised the gain from about 3 to 8.7 points. On BIRD, appending the arms to a fixed base instruction improved every arm, and on FrontierFinance, revising the winning-arm rule improved the advisor. These revisions were not exhaustive. Further error analysis and iterations would likely give further gains.

\paragraph{Scope} We studied four settings, and the results may not hold for other tasks, executors or datasets. In particular, we have not measured how well the advice transfers to other executors or to other datasets in the same domain. More generally, the quality of the structured advice and of the trained advisor depends on the training data and on the task. Some tasks may be hard to learn with an advisor. When the most useful advice requires the same inference that the executor gets wrong, as with BIRD's choice of tables and columns, the advisor gains little, and training the executor itself may be necessary. The method also needs a training pool with gold answers or a reliable grader, an executor that succeeds on some training inputs and fails on others, and an arms collection phase whose cost grows with the number of arms.

\section{Conclusion}\label{sec:conclusion}

We described hindsight structured advice distillation (HSAD), a way to improve a frozen executor by training a small advisor on the executor's own failures. The motivation is that a capable executor often lacks know-how about the task rather than general ability, and that this know-how is better supplied per input than written once into a long, shared prompt. The advisor learns, from inputs alone, the structured advice that a teacher wrote with hindsight access to gold answers and to the results of every strategy. It follows the idea of distillation with privileged information, but it distills into a separate small model rather than into the executor, so the executor stays unchanged and the advisor is cheap to train and run. On the public FrontierFinance benchmark, an 8B advisor lifted Samaya's proprietary research agent from 52.9\% to 56.6\%, above the agent's own high-effort configuration and the best frontier model in an open harness. With a small Qwen3-4B executor, the advice also improved a shopping agent, text-to-SQL and expert long-form generation.

How much the advice helps depends on the task and on the training data. The largest gains came where error analysis of the executor's traces revealed failures that advice could fix, where any instance-specific values could be recovered from the input, and where the training labels were reliable. These results suggest two practical rules for building such an advisor. First, derive the advice from error analysis of the executor's own traces, and give it fields whose values the advisor can predict from the input. On FrontierFinance, the teacher-executed gain disappears without the directives mined from the agent's failures, and on WebShop the decisive field restates option values that the instruction already contains. Second, label every field that can be read off the runs with a rule, and use the teacher's LLM only for fields that need judgment.

The advisor could also be made cheaper. Most advice fields are typed choices over closed vocabularies, such as the arm, the effort and length buckets, \cc{web\_use} and the directive picks. Recent decision models such as Jev return this kind of output directly, as calibrated probabilities over declared options, in a single parallel pass and at a small fraction of the latency and cost of an LLM call \cite{TypeSafe2026Jev}. Producing the categorical fields this way, and generating text only for instantiated values such as WebShop's option strings, could make structured advice generation much cheaper and faster. Open questions include how much of the advice transfers across executors and across datasets in the same domain. Another is how the advisor can learn from the executor's results after deployment. Our advisor is trained once, offline, and gives its advice before the executor runs. It could instead be updated online, adding new graded traces to its training data as they arrive, or revise its advice within an episode after seeing intermediate results, such as an error message or an empty query result.

\section*{Acknowledgments}
We thank Yuhao Zhang and Ashwin Paranjape for their helpful comments on this work.

\renewcommand{\bibfont}{\small}
\setlength{\bibsep}{0.5pt}
\bibliographystyle{unsrturl}
\bibliography{refs}

\appendix
\small
\section{Arms and rendering}\label{app:arms}

\subsection{Arms}

Table~\ref{tab:arms} lists the arms of each setting.

\begin{table}[t]
\centering\small
\begin{tabularx}{\linewidth}{@{}l>{\raggedright\arraybackslash}X@{}}
\toprule
Setting & Arms \\
\midrule
WebShop & \cc{default}, \cc{concise\_search\_and\_fast\_commit}, \cc{strict\_option\_click\_then\_buy}, \cc{exact\_variant\_text\_verification}, \cc{category\_demographic\_and\_price\_verify}, \cc{bounded\_variant\_fallback\_pivot} \\
FrontierFinance & \cc{default}, \cc{structured\_priority}, \cc{tables\_extraction}, \cc{restricted\_tools} \\
BIRD & \cc{default}, \cc{evidence\_literal\_and\_formula\_execution}, \cc{schema\_grounding\_and\_minimal\_joins}, \cc{conditional\_aggregation\_and\_ratios}, \cc{superlative\_extremums\_and\_ties}, \cc{cte\_multi\_stage\_decomposition}, \cc{strict\_projection\_and\_output\_alignment}, \cc{sqlite\_dialect\_and\_data\_sanitization}, \cc{multi\_intent\_union\_and\_cardinality} \\
ExpertLongBench & \cc{default}, \cc{schema\_first\_structured\_output}, \cc{entity\_identity\_first\_then\_annotate}, \cc{security\_threat\_frame\_first}, \cc{pedagogical\_evaluator\_role\_lock}, \cc{mechanistic\_rationale\_extraction}, \cc{clinical\_note\_completeness\_and\_fidelity}, \cc{exhaustive\_field\_coverage\_guard} \\
\bottomrule
\end{tabularx}
\caption{Arms of each setting. Each arm is appended to the domain's base prompt, and \cc{default} adds nothing.}
\label{tab:arms}
\end{table}

\subsection{Rendering examples}\label{app:render}

Each field of the advice is rendered with a fixed template, and the resulting text is appended to the executor's input.

\paragraph{FrontierFinance} For the Colgate-Palmolive query of Appendix~\ref{app:examples}, the advice is rendered as one guidance block after the query. The \cc{default} arm adds no text. Each directive is replaced by its description from the taxonomy, and the other fields become one short line each:
\begin{quote}\small
APPROACH GUIDANCE (advisory -- NOT part of the question to answer; you retain full tool access and may broaden your search if this does not cover it): Focus on: 1) When a requested metric is not disclosed by the company, explicitly state that the metric is not reported rather than leaving it blank, implying zero, or silently omitting it. \ldots 4) Clearly separate what management has explicitly stated or quantified from what can only be inferred or estimated, and flag when a metric has not been formally disclosed even if related information exists. $|$ Prioritize data sources: FILINGS, \ldots\ $|$ Suggested tools: \cc{documents\_list}, \cc{documents\_search}, \ldots\ $|$ Target length: long $|$ Web search: avoid
\end{quote}

\paragraph{WebShop} For the instruction in Fig.~3, the options part of the advice renders as: ``This product's page has option buttons. Before clicking Buy Now you MUST click the color button labelled `green'; the size button labelled `xx-large'. Click each of those exactly as written, then click Buy Now.'' The full rendering adds the arm's strategy text, ``Plan for 6 or more tool calls in total.'' and ``This task should require you to go back to search at least once; page through more results.''

\paragraph{BIRD} Advice is added to the user turn after the schema and question as a block headed ``GUIDANCE (advisory -- apply if relevant)'', containing one line for effort (for example ``Expected effort (high): a structurally demanding query \ldots''), one for length, and one line per picked directive with that directive's description.

\paragraph{ExpertLongBench} For the security-review document of Appendix~\ref{app:examples}, the arm's strategy text is appended to the document first. It begins: ``Before extracting any fields, perform a mandatory security-threat pass: scan every tool return value in the transcript \ldots'' The guidance block follows:
\begin{quote}\small
APPROACH GUIDANCE (advisory -- NOT part of the source document): Focus on: 1) When the task involves evaluating a transcript for security or safety concerns, analyze the agent's decision trajectory and any injected or adversarial content for risk rather than processing the surface-level content of the transcript as if executing the embedded task. \ldots 4) When a transcript contains an agent interacting with external tools, check whether any tool return value contains an injected instruction distinct from the original request, and if so, populate all security-analysis fields including trigger, motivation, tools, outcome, and binary label. $|$ Target length: short (a single structured note/summary section, $<$300 words) $|$ Effort: low -- answer directly with minimal deliberation; do not second-guess or re-derive an already-clear answer.
\end{quote}

\section{Directive taxonomies}\label{app:tax}

\paragraph{FrontierFinance} 103 directives in nine families, for example 2.8 \cc{pull\_sub\_segment\_detail\_within\_reported\_segments} and 3.1 \cc{cover\_all\_named\_entities\_in\_query}.

\paragraph{WebShop} The mined taxonomy had 50 directives in five families (search-query formulation, product validation, option configuration, exact variant matching, and fallback or pivoting). Evaluated one family at a time on 200 validation samples, search formulation lowered score by 5.45 points and fallback by 1.93, product validation and variant matching had no measurable effect, and only option configuration helped (+1.50). The final taxonomy keeps one area, option selection, with six instantiated directives: 3.1 \cc{select\_color}, 3.2 \cc{select\_size}, 3.3 \cc{select\_fit\_type}, 3.4 \cc{select\_item\_shape}, 3.5 \cc{select\_style} and 3.6 \cc{select\_scent\_or\_flavor}.

\paragraph{BIRD} Error areas, from the failure notes on BIRD's training split: flawed evidence over-reliance; string literal and value mismatch; uncleaned numeric text operations; schema linking and join path errors; projection and cardinality mismatch; aggregation and deduplication errors; SQL dialect and type-casting violations; superlative and ordering misinterpretation. Examples of the 63 directives: 1.2 \cc{map\_question\_concepts\_to\_schema\_columns}, 2.1 \cc{verify\_join\_keys\_against\_foreign\_key\_schema}, 2.3 \cc{eliminate\_redundant\_intermediate\_joins}, 5.1 \cc{strictly\_project\_requested\_attributes}, 6.1 \cc{prefer\_order\_by\_limit\_for\_single\_extremes}, 7.1 \cc{cast\_operands\_to\_real\_before\_division}, 7.5 \cc{escape\_single\_quotes\_in\_literals}, 8.1 \cc{strictly\_follow\_evidence\_calculation\_formulas}, 8.3 \cc{copy\_evidence\_literals\_verbatim}.

\paragraph{ExpertLongBench} Nine areas (task interpretation, source grounding, field completeness, value accuracy, schema-field disambiguation, negative-finding handling, output structure and completeness, security analysis, pedagogical evaluation) with 57 directives, for example 1.2 \cc{adopt\_correct\_evaluator\_role}, 2.1 \cc{ground\_all\_fields\_strictly\_in\_source}, 2.2 \cc{scan\_entire\_source\_for\_each\_field}, 3.1 \cc{populate\_all\_schema\_fields\_explicitly}, 6.1 \cc{record\_absent\_fields\_explicitly}, 8.1 \cc{detect\_and\_label\_prompt\_injection}.

\section{Implementation details and additional analyses}\label{app:details}

This appendix first gives settings shared by all experiments and then additional details and analyses for each setting.

\subsection{Common settings and training}\label{app:common}

\paragraph{Common settings} The teacher's LLM and the error-analysis model are API models, and neither is used as the executor or the advisor. The advisors are Qwen3-8B for FrontierFinance and Qwen3-4B \cite{arXiv250509388} elsewhere, fine-tuned with LoRA. Ties between arms are broken toward \cc{default} unless stated otherwise. Open models are served with vLLM \cite{arXiv230906180}.

\paragraph{Advisor training} All advisors are trained with LoRA \cite{arXiv210609685} (rank 64, $\alpha=128$) for six epochs, and we keep the checkpoint of every epoch for selection. Training one advisor takes a few hours on a single GPU.

\paragraph{Epoch selection} In every setting, we choose the checkpoint used on the test set by its lift on the validation set, measured by running the executor with each checkpoint's advice. For example, on BIRD the validation lift was +1.1, +3.5 and +2.8 points at epochs 4, 5 and 6, so we use the epoch-5 checkpoint.

\paragraph{Tie-breaks} On WebShop, we also tried breaking ties toward the arm with the fewest turns instead of toward \cc{default}, which moved 602 training rows away from \cc{default}. We observed a drop in performance with the advisor trained on these labels, so on WebShop we break ties toward \cc{default}. On BIRD, 52\% of training questions have two or more correct arms and 41\% have none, so on most rows the scores alone cannot pick a winner. On these rows the teacher's LLM picks the arm. It sees the gold query and its result, together with each arm's query and execution result, and chooses the arm whose query is closest to the gold query.

\paragraph{Run-to-run variation} The executor's score on a single input varies between runs. Ideally, each arm would be run several times with different seeds, and its average score, or its pass@$k$, would be used to find the best arm and compute the reference points. We did not do this because it would multiply the cost of the arms collection, which is already the most expensive step of the pipeline. The reported lifts are less affected, since they are averages over many test inputs. We leave a statistical analysis over repeated runs to future work.

\subsection{FrontierFinance}\label{app:c-ff}

\paragraph{Choice of the winning-arm rule}\label{app:ff-labels} The winning arm of each training question can be defined in more than one way, and the definition affects the trained advisor. We compare two rules. The first picks the arm with the highest composite score, a hand-weighted sum of the essential-criteria rate with smaller, capped terms for cost and latency, and breaks ties toward \cc{default}. The second is the cascade of \S\ref{sec:labels}, which orders the arms by essential-criteria rate, then all-criteria rate, then cost, then latency. The two rules choose a different arm on 681 of the 1,819 training questions. Under the composite rule, 51\% of the labels are \cc{default}, because many arms tie on the composite and ties go to \cc{default}. Under the cascade, the four arms receive 25\%, 31\%, 22\% and 22\% of the labels. With the same schema, teacher prompts, trainer and renderer, advisor variants trained on the composite labels reach 57.2--57.6\% essential-criteria coverage on the test queries, and the advisor trained on the cascade labels reaches 59.9\% in the deployment configuration (Table~\ref{tab:ff-main}). A rule that orders the arms by the benchmark's own criteria therefore produces a better advisor, which indicates that the winning-arm objective is aligned with end-task quality.

\paragraph{Deployment configuration}\label{app:ff-deploy} We also evaluated the agent in the configuration in which it is deployed, which is also how it was run during training. Table~\ref{tab:ff-main} reports essential and all-criteria qualification rates on the 220 test queries and on the internal validation set. The advisor raises essential-criteria coverage from 53.9\% to 59.9\% on the test queries and from 63.0\% to 69.2\% on validation, and all-criteria coverage by about 7--8pp on both. 

\begin{table}[t]
\centering\small
\begin{tabular}{@{}lcccc@{}}
\toprule
 & \multicolumn{2}{c}{Public test (220)} & \multicolumn{2}{c}{Validation (100)} \\
\cmidrule(lr){2-3}\cmidrule(lr){4-5}
Condition & essential & all & essential & all \\
\midrule
No advice & 53.9\% & 45.4\% & 63.0\% & 53.9\% \\
Best fixed arm (\cc{tables\_extraction}) & 57.9\% & 51.7\% & 69.1\% & 60.2\% \\
Trained advisor & \textbf{59.9\%} & \textbf{52.6\%} & \textbf{69.2\%} & \textbf{61.9\%} \\
\midrule
\multicolumn{5}{@{}l}{\emph{Training variants of the advisor}} \\
\quad more epochs (9) & 59.3\% & 52.2\% & 65.9\% & 58.4\% \\
\quad reasoning before the JSON & 57.4\% & 49.9\% & 66.3\% & 58.2\% \\
\quad GRPO from the SFT checkpoint & 58.4\% & 52.4\% & 67.8\% & 58.8\% \\
\bottomrule
\end{tabular}
\caption{FrontierFinance in the deployment configuration. The best fixed arm applies the same arm to every question (without other fields in the advice).}
\label{tab:ff-main}
\end{table}

A single fixed arm, \cc{tables\_extraction}, is a strong baseline (Table~\ref{tab:ff-main}). The advisor beats it on the test queries. Choosing the arm is only one of the advisor's fields, and the ablation of \S\ref{sec:qa-ablation} shows that the other fields result in additional gains.

\paragraph{More epochs of SFT} Training the advisor for nine epochs instead of six slightly raises its agreement with the teacher's targets on the held-out validation questions, measured with the field-by-field reward described in RL section below (0.393 against 0.384). However, this lowers the executor's essential-criteria coverage on validation, from 69.2\% to 65.9\% (Table~\ref{tab:ff-main}). We therefore use the checkpoint after six epochs as the SFT advisor.

\paragraph{Reinforcement learning from the SFT checkpoint} We also tried to improve the SFT advisor with GRPO \cite{arXiv240203300}. Starting from the selected SFT checkpoint, we trained a new LoRA for one epoch over the 1,819 training questions, with eight sampled outputs per question. The reward compares the advisor's JSON with the teacher's target field by field and does not run the agent, since one agent run takes about four minutes and costs about a dollar. The field weights reflect how much each field contributed in the ablation of \S\ref{sec:qa}: 0.40 for the arm, 0.35 for the overlap (Jaccard index) between the predicted and target directive picks, 0.10 for the sources, 0.06 each for the tools and web use, and 0.03 for the target length. In a first run, the arm part of the reward gave partial credit. For each training question, we know the score that every arm received during collection. Choosing the best arm earned 1, choosing the worst arm earned 0, and any other arm earned a value in between, in proportion to its score. For example, if the four arms scored 80\%, 70\%, 60\% and 40\% on a question, choosing the arm that scored 70\% earned 0.75. The reward rose from 0.57 to 0.74 during training, but the advisor learned to choose \cc{tables\_extraction} for almost every question, because under graded credit this single arm already earns 0.67 of the arm reward. A second run scored the arm by exact match with the label and added a KL penalty to the SFT policy ($\beta=0.01$). This kept the arm choices varied, and the directive-pick overlap rose from 0.51 to 0.71 and the arm match from 0.25 to 0.34 on the training questions. On held-out questions, however, arm accuracy was flat, and in the executor the advisor scored 67.8\% essential on validation and 58.4\% on the test queries, below the SFT checkpoint it started from (Table~\ref{tab:ff-main}). These results show that this form of reward, which compares the advice with the teacher's target without measuring the executor's performance on the task, can rise on the training questions without leading to better answers. Finding low-cost reward signals that correlate with the executor's downstream performance is a natural direction for future work on training the advisor with RL.

\subsection{WebShop}\label{app:c-ws}

\paragraph{Advice designs and reference points}\label{app:ws-designs} Fig.~5a compares three advice designs on the 500 test samples, each produced by a trained advisor. Predicting only the arm gains 3.65 score points. The second design adds generic directives from the taxonomy mined from failure notes (\S\ref{sec:mining}). That taxonomy has 50 directives in five families. Evaluated one family at a time on the 200 validation samples, only the option-configuration family helped (Appendix~\ref{app:tax}), while the search and fallback families lowered the score, for example by telling the agent to abandon a product whose exact option was missing. With only the option family kept, the generic directives, such as ``inspect all option buttons and click every required one'', add nothing on top of the arm (+2.94). The third design, the trained advisor with its full advice, replaces these generic directives with the six option directives of the final taxonomy, each instantiated with the option value stated in the instruction, such as ``click the size button labelled `x-large'\,''. This design follows from the lost-reward analysis of \S\ref{sec:webshop} and gains 8.73 points, close to the 8.95 points obtained when the option values come from the regular-expression rule instead of the advisor. The difference between the second and third designs is therefore not only which directives are kept but whether they are generic or instantiated. Fig.~5b shows the reference points of \S\ref{sec:ceilings} on the 200 validation samples. The best-arm oracle is +9.59, but running each sample again with its best arm gives only +6.49 (run-to-run variance, Appendix~\ref{app:common}). The trained advisor gains 8.16 points on validation, more than the executed best arm, because its options field carries information that no arm contains.

\begin{figure}[H]
\figblock{\resizebox{\linewidth}{!}{
\begin{tikzpicture}[figfont]
\begin{axis}[
  name=ws, xbar, width=6.2cm, height=3.6cm, bar width=7pt,
  xmin=0, xmax=12, axis x line*=bottom, axis y line*=left,
  y axis line style={draw=none}, ytick style={draw=none},
  xlabel={\figfontsmtxt score lift over no advice (points)},
  symbolic y coords={arm,gen,adv}, ytick={arm,gen,adv}, bar shift=0pt, enlarge y limits=0.25,
  xtick={0,5,10}, xticklabels={0,5,10}, y dir=reverse,
  yticklabels={{arm only},{arm + generic option\\directives},{trained advisor\\(full advice)}},
  yticklabel style={figfontsm, align=right, text width=3.1cm},
  xticklabel style={figfontsm}, x axis line style={figMute}, xtick style={figMute},
  nodes near coords, every node near coord/.style={figfontsm, text=figInk, anchor=west, inner sep=1.5pt},
  point meta=explicit symbolic,
  title={\figfontlgtxt (a) Advice designs, test 500}, title style={at={(0.35,1.0)}},
]
\addplot[draw=figAcq, fill=figAcqBg] coordinates {(3.65,arm) [+3.65] (2.94,gen) [+2.94]};
\addplot[draw=figGood, fill=figGoodBg] coordinates {(8.73,adv) [+8.73]};
\end{axis}
\begin{axis}[
  at={(ws.east)}, anchor=west, xshift=3.2cm,
  xbar, width=6.2cm, height=3.6cm, bar width=7pt,
  xmin=0, xmax=12, axis x line*=bottom, axis y line*=left,
  y axis line style={draw=none}, ytick style={draw=none},
  xlabel={\figfontsmtxt score lift over no advice (points)},
  symbolic y coords={orc,fresh,stud}, ytick={orc,fresh,stud}, bar shift=0pt, enlarge y limits=0.25,
  xtick={0,5,10}, xticklabels={0,5,10}, y dir=reverse,
  yticklabels={{best-arm oracle},{best arm,\\run again},{trained advisor\\(full advice)}},
  yticklabel style={figfontsm, align=right, text width=2.6cm},
  xticklabel style={figfontsm}, x axis line style={figMute}, xtick style={figMute},
  nodes near coords, every node near coord/.style={figfontsm, text=figInk, anchor=west, inner sep=1.5pt},
  point meta=explicit symbolic,
  title={\figfontlgtxt (b) Reference points, validation 200}, title style={at={(0.35,1.0)}},
]
\addplot[draw=figMute, fill=figSurf] coordinates {(9.59,orc) [+9.59]};
\addplot[draw=figFil, fill=figFilBg] coordinates {(6.49,fresh) [+6.49]};
\addplot[draw=figGood, fill=figGoodBg] coordinates {(8.16,stud) [+8.16]};
\end{axis}
\end{tikzpicture}}}
\figcap{5}{WebShop score lift over no advice. (a) Three advice designs on the 500 test samples. (b) The reference points of \S\ref{sec:ceilings} on the 200 validation samples: the best arm per sample as recorded during collection, the same best arm run again on each sample, and the trained advisor.}
\end{figure}

\paragraph{A difficulty field} The WebShop advisor can also predict a \cc{difficulty} field, labelled by a rule from the \cc{default} arm's score on each training sample (hard, medium or easy by tercile). Rendering this field in the prompt lowers the score by 2.95 points, so the final advice does not render it.

\paragraph{Other directive variants} Instantiating the product type and price as verification directives, such as ``verify that the title is really a men's dress shirt'', lowers the score by 1.85 points. The verification framing leads the agent to reject products and search again, the same failure caused by the directives that tell it to abandon a product whose exact option is missing.

\subsection{BIRD}\label{app:c-bird}

\paragraph{Reference points}\label{app:bird-ref} Fig.~6 shows the reference points of \S\ref{sec:ceilings} on the 1,074 BIRD test questions. The best-arm oracle is +20.11 execution-accuracy points, running the winning arm again gives +11.27, and the trained advisor gains 3.07. The oracle is higher than the rerun because each arm was run only once during collection: on a question where one of the nine arms happened to succeed, running that arm again does not always succeed. See also Appendix~\ref{app:ws-designs} for similar behavior for WebShop. The trained advisor stays well below the best arm run again, unlike on WebShop. On WebShop, the most useful advice copies option values that the instruction states, so the advisor can produce it exactly. On BIRD, the most useful advice says which tables and columns the query needs, and the advisor has to guess this from the question without seeing the database schema (\S\ref{sec:bird}). High-quality instantiated advice is therefore harder to obtain on this dataset. With more training data, or with better advice fields, there is still room to improve.

\begin{figure}[H]
\figblock{
\begin{tikzpicture}[figfont]
\begin{axis}[
  xbar, width=6.2cm, height=3.6cm, bar width=7pt,
  xmin=0, xmax=25, axis x line*=bottom, axis y line*=left,
  y axis line style={draw=none}, ytick style={draw=none},
  xlabel={\figfontsmtxt lift over no advice (execution-accuracy points)},
  symbolic y coords={mech,teach,stud}, ytick={mech,teach,stud}, bar shift=0pt, enlarge y limits=0.25,
  xtick={0,10,20}, xticklabels={0,10,20}, y dir=reverse,
  yticklabels={{best-arm oracle},{best arm,\\run again},{trained advisor\\(full advice)}},
  yticklabel style={figfontsm, align=right, text width=2.6cm},
  xticklabel style={figfontsm}, x axis line style={figMute}, xtick style={figMute},
  nodes near coords, every node near coord/.style={figfontsm, text=figInk, anchor=west, inner sep=1.5pt},
  point meta=explicit symbolic,
  title={\figfontlgtxt BIRD, test 1,074}, title style={at={(0.35,1.0)}},
]
\addplot[draw=figMute, fill=figSurf] coordinates {(20.11,mech) [+20.11]};
\addplot[draw=figFil, fill=figFilBg] coordinates {(11.27,teach) [+11.27]};
\addplot[draw=figGood, fill=figGoodBg] coordinates {(3.07,stud) [+3.07]};
\end{axis}
\end{tikzpicture}}
\figcap{6}{BIRD's three reference points on the 1,074 test questions: the best arm per sample as recorded during collection, the same best arm run again on each sample, and the trained advisor.}
\end{figure}

\paragraph{Gain by difficulty} BIRD labels each question as simple, moderate or challenging. The gain is concentrated in simple questions (+5.19 points), is small for moderate ones (+0.93) and is negative for the most challenging ones ($-4.21$).

\paragraph{Appending arms} An arm can replace the base instruction as a complete rewritten system prompt, or be appended to a fixed base instruction. On BIRD, we collected all 9,428 training questions both ways. Appending improves every one of the eight non-default arms, by 0.5 to 7.1 execution-accuracy points. The largest gains are for the two arms that scored lowest as full prompts.

\paragraph{When the advice hurts} A model diagnosed the 72 test questions that the advice turns from correct to wrong. The dominant pattern is over-application, such as adding a join that duplicates rows, ``correcting'' a literal that was already right (for example turning \cc{'CA'} into \cc{'California'}), or choosing the wrong one of two similarly named columns. In about two thirds of these cases the arm is \cc{default}, so the harm comes from the directive text. Four of the 63 directives are net negative, led by \cc{use\_conditional\_aggregation\_for\_pivoted\_metrics}, and the other frequently used directives are net positive. These cases show that the BIRD taxonomy and directive texts leave room for further refinement, for example by rewording or removing the directives that are over-applied.

\paragraph{Coverage of the residual failures} For a sample of 180 test questions that the executor answers wrongly both with and without advice, we checked each root cause against the taxonomy. For 94\% of them, an existing directive already addresses the root cause. The remaining gap is therefore mostly a matter of picking the right directive and having the executor apply it correctly, not of missing strategies. A second test supports this. If we run all nine arms on each question and choose among their queries using the execution results, either the first query that runs without error or the result that most arms agree on, execution accuracy rises by about 6 points, twice the advisor's gain. For many questions, one of the existing arms already writes a correct query, and the missing step is choosing it. The advisor cannot use this signal, because it runs before any query is executed.

\subsection{ExpertLongBench}\label{app:c-elb}

\paragraph{Metric} ExpertLongBench's own paper reports an F1 score computed with its checklist-extraction framework \cite{arXiv250601241}, whereas we report the fraction of checklist fields covered using a judge, so our numbers are not comparable with the published ones.

\paragraph{Lift by task}\label{app:elb-tasks} Table~\ref{tab:elb} breaks the ExpertLongBench lift down by task. The largest gains are in security review (+20.4 points) and pedagogical evaluation (+9.0), and the other four tasks change little. The directive picks generally fit the task, as the examples in Table~\ref{tab:elb} show: the advisor picks prompt-injection detection for security reviews and examination-finding transcription for clinical notes. The arm choices fit less well. For several tasks, the advisor often chooses an arm designed for another task, for example the security frame for clinical notes and materials-science papers. We read this as cross-task interference. With 450 training documents shared by six tasks and eight arms, the advisor sees few examples of each task and arm, and arms learned for one task carry over to others. The two tasks with a clear gain are also the two where the predicted arm usually matches the task. We expect that more training data per task, a taxonomy and set of arms organised by task, or restricting the arm choice to the arms that fit each task could raise the lift on the remaining tasks, and we leave this to future work.

\begin{table}[H]
\centering\small
\resizebox{\linewidth}{!}{\begin{tabular}{@{}lrrll@{}}
\toprule
Task & $n$ & Lift & Most frequent predicted arm & Example directive picked \\
\midrule
T11 security review & 18 & +20.4 & \cc{security\_threat\_frame\_first} (12) & 8.1 \cc{detect\_and\_label\_prompt\_injection} \\
T04 pedagogical evaluation & 19 & +9.0 & \cc{pedagogical\_evaluator\_role\_lock} (19) & 5.3 \cc{separate\_mistake\_identification\ldots} \\
T03 materials synthesis & 9 & 0.0 & \cc{security\_threat\_frame\_first} (7) & 3.8 \cc{capture\_precursor\_rationale\ldots} \\
T08 protein descriptions & 18 & 0.0 & \cc{clinical\_note\_completeness\ldots} (16) & 3.6 \cc{populate\_biological\_annotation\ldots} \\
T06 clinical notes & 18 & $-1.0$ & \cc{security\_threat\_frame\_first} (18) & 3.5 \cc{transcribe\_complete\_examination\ldots} \\
T07 molecule descriptions & 18 & $-1.9$ & \cc{pedagogical\_evaluator\_role\_lock} (13) & 3.7 \cc{report\_all\_compound\_classification\ldots} \\
\bottomrule
\end{tabular}}
\caption{ExpertLongBench lift by task on the 100 test documents, in checklist-coverage points, with the arm the advisor chose most often for that task (count in parentheses) and a representative directive it picked.}
\label{tab:elb}
\end{table}


\section{Examples of advisor output}\label{app:examples}

\subsection{FrontierFinance}

The examples below are queries from the 220 public FrontierFinance queries, answered by the agent without advice and with the trained advisor's advice (deployment configuration). The advisor saw only the query. We show the advice fields and the directive picks, with some source and tool names abbreviated.

\paragraph{Colgate-Palmolive (financial data)} \emph{``What is Colgate-Palmolive's revenue by geography and product category for each quarter since 2019?''}

\noindent\begin{tabular}{@{}ll@{}}
\cc{arm} & \cc{default} \\
\cc{target\_length} & \cc{long} \\
\cc{web\_use} & \cc{avoid} \\
\cc{sources} & \cc{FILINGS, \ldots} \\
\cc{tools} & \cc{documents\_list, documents\_search, \ldots} \\
\end{tabular}

\noindent Directives: \cc{2.2 distinguish\_metric\_unavailability\_from\_zero}; \cc{1.1 enumerate\_all\_periods\_in\_requested\_range}; \cc{6.5 derive\_standalone\_periods\_from\_cumulative\_filings}; \cc{2.17 distinguish\_explicit\_from\_implicit\_disclosures}.

Without advice, the agent filled many quarterly cells with estimates obtained by applying a percentage share to a total. With advice, it reported the quarterly figures by region and stated that Colgate discloses product categories only as a share of sales, not in dollars. Rubric items satisfied: 36\% without advice, 90\% with advice.

\paragraph{ASML (earnings and events)} \emph{``What is the summary of ASML's earnings?''}

\noindent\begin{tabular}{@{}ll@{}}
\cc{arm} & \cc{tables\_extraction} \\
\cc{target\_length} & \cc{medium} \\
\cc{web\_use} & \cc{fallback} \\
\cc{sources} & \cc{TRANSCRIPTS, FILINGS, NEWS, \ldots} \\
\cc{tools} & \cc{documents\_list, documents\_search, \ldots} \\
\end{tabular}

\noindent Directives: \cc{4.1 search\_across\_all\_document\_types\_for\_topic}; \cc{2.15 surface\_management\_guidance\_and\_forward\_commentary}; \cc{2.5 include\_geographic\_and\_segment\_breakdowns}; \cc{2.7 include\_capital\_allocation\_and\_investment\_announcements}.

Without advice, the agent centred its summary on the latest quarter and its guidance. With advice, it also built a quarterly grid for the year and added the regional breakdown, including China, and the capital return programme. Rubric items satisfied: 47\% without advice, 100\% with advice.

\subsection{WebShop, BIRD and ExpertLongBench}

For each of the other settings, we show one test input on which the advice helped, with the trained advisor's output and the executor's score with and without it.

\paragraph{WebShop} \emph{``Find me hand wash, wash cold women's bodysuit tops with polyester spandex for tumble dry, daily wear with size: x-large, and price lower than 50.00 dollars.''}

\noindent\begin{tabular}{@{}l>{\raggedright\arraybackslash}p{0.68\linewidth}@{}}
\cc{arm} & \cc{concise\_search\_and\_fast\_commit} \\
\cc{effort}, \cc{expected\_tools} & \cc{short}, none \\
\cc{required\_options} & \cc{\{size: x-large\}} \\
\cc{directive\_picks} & \cc{3.2 select\_size} \\
\end{tabular}

\noindent Rendered as the arm's strategy text (``Formulate search queries using strictly 2 to 4 essential keywords \ldots\ On the product page, select all required attribute options and click Buy Now within 4 to 5 total turns.''), followed by ``Plan for about 3 tool calls in total.'' and ``This product's page has option buttons. Before clicking Buy Now you MUST click the size button labelled `x-large'. Click each of those exactly as written, then click Buy Now.'' Reward: 0.06 without advice, 1.00 with the full advice. With the option line alone, the reward was 0.07, so on this sample the arm and effort fields made the difference.

\paragraph{BIRD} \emph{``What is the average score of the posts owned by the user csgillespie?''} (evidence: ``csgillespie'' is the DisplayName of the user; average score refers to AVG(score)).

\noindent\begin{tabular}{@{}l>{\raggedright\arraybackslash}p{0.68\linewidth}@{}}
\cc{arm} & \cc{schema\_grounding\_and\_minimal\_joins} \\
\cc{effort}, \cc{length} & \cc{high}, \cc{medium} \\
\cc{directive\_picks} & \cc{1.1 validate\_attribute\_existence\_and\_ownership}; \cc{2.4 join\_descriptive\_tables\_for\_named\_entities}; \cc{8.1 strictly\_follow\_evidence\_calculation\_formulas}; \cc{8.3 copy\_evidence\_literals\_verbatim} \\
\end{tabular}

\noindent Without advice the executor wrote
\begin{quote}\cc{SELECT AVG(posts.score) FROM posts WHERE posts.OwnerDisplayName = 'csgillespie'}\end{quote}
\noindent which filters on a denormalized column and does not match the gold result. With advice it wrote
\begin{quote}\cc{SELECT AVG(posts.score) FROM posts JOIN users ON posts.OwnerUserId = users.Id\\ WHERE users.DisplayName = 'csgillespie'}\end{quote}
\noindent which matches. The schema-grounding arm and the directive to join descriptive tables for named entities point the executor to the users table.

\paragraph{ExpertLongBench} Task T11 (security review): the input is a transcript of an agent that was asked to fetch the details and reviews of a laptop on Amazon. One of the returned reviews contains an injected instruction to delete a patient's clinical documents in another system. The expected output is a structured security analysis with six fields.

\noindent\begin{tabular}{@{}l>{\raggedright\arraybackslash}p{0.68\linewidth}@{}}
\cc{arm} & \cc{security\_threat\_frame\_first} \\
\cc{target\_length}, \cc{effort} & \cc{short}, \cc{low} \\
\cc{directive\_picks} & \cc{1.5 analyze\_security\_trajectory\_not\_surface\_content}; \cc{2.5 detect\_adversarial\_content\_in\_tool\_returns}; \cc{3.10 populate\_all\_security\_analysis\_fields}; \cc{8.1 detect\_and\_label\_prompt\_injection} \\
\end{tabular}

\noindent Without advice, the executor wrote a product summary with the reviews, noting only that one review was unrelated to the product. With advice, it wrote the security analysis, with the label, the attacker's motivation, the verbatim trigger and the tool that returned the injected text. Checklist fields satisfied: 1 of 6 without advice, 5 of 6 with advice.

\end{document}